\documentclass[journal]{IEEEtran}
\usepackage{amsmath,amsfonts}
\usepackage{algorithm2e}
\usepackage{array}
\usepackage{textcomp}
\usepackage{stfloats}
\usepackage{url}
\usepackage{verbatim}
\usepackage{graphicx}
\usepackage{adjustbox}
\usepackage{cite}
\usepackage{hyperref}
\usepackage{url}
\usepackage{booktabs}
\usepackage{braket} 
\usepackage{soul}
\usepackage{placeins}
\usepackage[dvipsnames]{xcolor}
\usepackage{colortbl} 

\SetKwComment{Comment}{/* }{ */}
\RestyleAlgo{ruled}

\usepackage{xcolor} 

\usepackage[switch]{lineno}
\usepackage{multicol}
\usepackage{multirow}
\usepackage{mwe}
\usepackage{float}
\usepackage{balance}
\usepackage{lipsum}
\usepackage{balance}

\usepackage{amsmath,amsfonts}
\usepackage{algorithm2e}
\usepackage{array}
\usepackage{textcomp}
\usepackage{stfloats}
\usepackage{url}
\usepackage{verbatim}
\usepackage{graphicx}
\usepackage{cite}
\usepackage{hyperref}
\usepackage{url}
\usepackage{booktabs}
\usepackage{braket} 
\usepackage[dvipsnames]{xcolor}
\usepackage{flushend}

\usepackage{xcolor}
\usepackage{soul}

\usepackage{multicol}
\usepackage{multirow}
\usepackage{diagbox}

\usepackage{mwe}
\usepackage{float}
\usepackage{balance}
\usepackage{tikz}
\usetikzlibrary{quantikz2}
\usepackage{xfrac}
\usepackage{amssymb}

\ifCLASSINFOpdf
\else
\fi
\begin{document}
%
\title{A Deep Learning Architecture for Land Cover Mapping Using Spatio-Temporal Sentinel-1 Features}
%
%
%

\author{\IEEEauthorblockN{L. Russo, \IEEEmembership{Student Member, IEEE},
A. Sorriso, \IEEEmembership{Member, IEEE}, 
S. L. Ullo, \IEEEmembership{Senior Member, IEEE}, 
and \\P. Gamba, \IEEEmembership{Fellow, IEEE} 
}
\thanks{
\hspace{-0.4cm}
\textit{(Corresponding author: Antonietta Sorriso)}\\
Luigi Russo, Antonietta Sorriso and Paolo Gamba are with the Department of Electrical, Computer and Biomedical Engineering, University of Pavia, 27100 Pavia, Italy (e-mail: luigi.russo02@universitadipavia.it; antonietta.sorriso@unipv.it; paolo.gamba@universitadipavia.it). \\
Silvia Liberata Ullo is with the Department of Engineering, University of Sannio, 82100 Benevento, Italy (email: ullo@unisannio.it).\\
This work was supported by ESA in the framework of ESA CCI+ High Resolution (HR)
LC project, Phase 2 (https://climate.esa.int/en/projects/high-resolution-land-cover/).}.}

\maketitle
\begin{abstract}
Land Cover (LC) mapping using satellite imagery is critical for environmental monitoring and management. Deep Learning (DL), particularly Convolutional Neural Networks (CNNs) and Vision Transformers (ViTs), have revolutionized this field by enhancing the accuracy of classification tasks. 
In this work, a novel approach combining a transformer-based Swin-Unet architecture with seasonal synthesized spatio-temporal images has been employed to classify LC types using spatio-temporal features extracted from Sentinel-1 (S1) Synthetic Aperture Radar (SAR) data, organized into seasonal clusters.
The study focuses on three distinct regions - Amazonia, Africa, and Siberia - and evaluates the model performance across diverse ecoregions within these areas. By utilizing seasonal feature sequences instead of dense temporal sequences, notable performance improvements have been achieved, especially in regions with temporal data gaps like Siberia, where S1 data distribution is uneven and non-uniform. The results demonstrate the effectiveness and the generalization capabilities of the proposed methodology in achieving high overall accuracy (O.A.) values, even in regions with limited training data.
\end{abstract}


%
\IEEEpeerreviewmaketitle

\section{Introduction}
\label{intro}
%
%
%
%

\IEEEPARstart{L}{and} cover  (LC) mapping has emerged as a critical tool for a diverse range of applications including forest monitoring, agriculture, urbanization, flood monitoring, and climate change analysis. Accurate LC maps are essential for shaping effective land use policies. The LC maps are extremely beneficial for evaluating ecosystem health. They allow users to monitor conditions across different regions worldwide, aiding in more informed decisions for effective ecosystem management. Additionally, these maps help track environmental health and identify the impacts of ongoing changes.

The advent of free optical and SAR datasets, such as those from the Sentinel constellation, combined with cloud processing platforms like Google Earth Engine (GEE) \cite{gorelick2017google} and the Copernicus Data and Information Access Services (DIAS) platforms, has greatly facilitated LC mapping on a broad scale.\\
Over the years, a variety of methodologies have been explored for LC mapping. Early approaches primarily utilized optical imagery, as seen in works that leveraged the Landsat archive for Eastern Europe and Romania \cite{6415303, 9831397}. More recent studies have begun incorporating SAR data, often in combination with advanced machine learning (ML) techniques, to enhance the accuracy and detail of LC maps \cite{dahhani2023land}, \cite{waske2009classifier}, \cite{deluca2022integrated}, \cite{marzi2023automatic}. In the latter, for instance,  a ML-based approach using a Random Forest (RF) classifier on multitemporal SAR Sentinel-1 (S1) data to classify vegetation LC types is employed, highlighting the ongoing evolution in methodology and application scope.

The integration of DL methodologies, particularly Convolutional Neural Networks (CNNs), has marked a significant advancement in the field of image processing, including remote sensing for LC mapping. CNNs are exceptionally adept at extracting local spatial patterns directly from raw inputs through their convolutional layers, thereby learning enhanced feature representations \cite{zhang2016deep}, \cite{Lu2007Survey}, \cite{obj_det}, \cite{Hao2020Brief}. Their encoder-decoder architectures, exemplified by models like U-Net \cite{ronneberger2015unet}, efficiently map satellite imagery into detailed segmentation maps while preserving spatial details through skip connections.
However, the emergence of Vision Transformers (ViTs) has introduced a new paradigm in understanding global dependencies within images. Unlike CNNs, ViTs utilize multihead attention mechanisms to capture long-range contextual relationships, offering a distinct advantage in scenarios where global contextual understanding is crucial \cite{dosovitskiy2021image}.
In the specific context of SAR data for LC mapping, several DL architectures have been tailored to address the unique challenges presented by this type of data. CNNs remain a popular choice, as evidenced by their extensive use in various studies. For example, \cite{7891032} uses a CNN architecture applied in a heterogeneous environment for crop classification in the Kyiv region of Ukraine, using time series acquired by Landsat-8 and S1A, achieving an Overall Accuracy (O.A.) of about 94\%. \cite{Fontanelli2022} evaluated the performance of COSMO-SkyMed X-band dual-polarized data in the test area in Ponte a Elsa (Tuscany, Central Italy) in January–September 2020 and 2021. In this case, a CNN based classifier was arranged, trained and used for the LC mapping, and an O.A. above 90\% was marked.

Other innovative approaches include the integration of Recurrent Neural Networks (RNNs) to better handle the temporal dynamics of multitemporal SAR data, enhancing classification accuracy significantly over traditional ML methods \cite{ndikumana2018deep}.

Additionally, hybrid models that combine CNNs with RNNs, such as the Fully Convolutional Network (FCN) combined with Convolutional Long Term Memory (ConvLSTM), have demonstrated their effectiveness in extracting both spatial and temporal features from SAR data \cite{teimouri2019novel}. Such integrative approaches exemplify the ongoing innovation in DL techniques tailored for enhanced LC mapping using SAR data.

In summary, most traditional high-resolution (HR) LC maps are generated using either optical data alone or a combination of SAR and optical data, and are predominantly focused on limited regional areas \cite{bartsch2016land}, \cite{inglada2017operational}, \cite{buchhorn2020copernicus} while only a few provide global-scale coverage, relying solely on optical sources \cite{venter2022global}. In contrast, medium-resolution maps, with spatial resolutions of 300m \cite{bontemps2015multi} or 100m \cite{buchhorn2020copernicus} , offer broader global coverage. 
The exclusive use of SAR data, combined with advanced DL techniques such as CNN and ViT, has enabled more comprehensive and detailed analyses of LC on a global scale. This evolution underscores the dynamic nature of remote sensing technologies and their growing importance in global environmental monitoring and policy-making.
Building upon the significant progress made in the application of DL and CNNs for LC mapping using SAR data, the approach delineated in this work uniquely advances the field through the strategic utilization of synthesized spatio-temporal images, also called \textit{features}, on large scale. The features are synthetic images which well exploit the SAR time series to extract the temporal and spatial information concerning the observed "real world". These synthetic images, the features, are obtained by pure computation, i.e. by modelling the real world through the application of the temporal and spatial filters.

\begin{figure*} [htp]
	\centering
	\adjustbox{max width=\textwidth}{\includegraphics[]{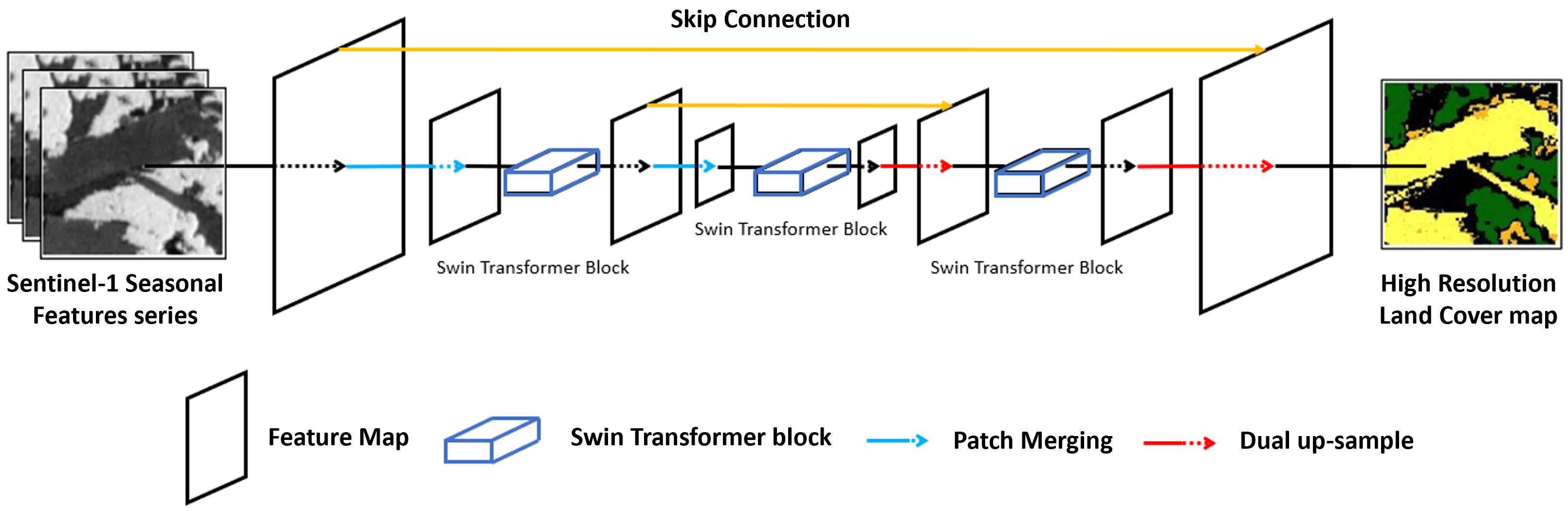}}
	\caption{Overview of the employed Swin-UNet architecture described in Section \ref{sec:swin_unet} for the LC classification task. Distinctive aspects of the Swin Transformer architecture are combined with the UNet architecture to achieve optimal performance.}
	\label{fig:swin_workflow}
\end{figure*}

The existing literature predominantly focuses on satellite-based mapping solutions that rely heavily on optical imagery, which is significantly constrained by weather conditions. In contrast, studies utilizing dense and extended temporal series of SAR imagery for LC mapping remain relatively scarce. This work introduces a DL-based approach that leverages radar-derived information, incorporating both temporal and spatial features extracted from S1 imagery. These features are systematically organized into seasonal clusters rather than relying on continuous dense sequences. This methodology stands out by providing not only spatial information but also integrating temporal data through a novel seasonal division approach. This strategy markedly enhances LC classification performance, particularly in scenarios where Sentinel-2 (S2) tiles exhibit sparse temporal coverage. For instance, \cite{ghassemi2024european} uses a combination of features derived from S2 and S1 data. For S1, the features analyzed include temporal monthly and yearly statistical measures - such as the median, 5th, 50th, and 98th percentiles - calculated from the VH and VV polarimetric bands, as well as their combination (ratio and difference), and various indices ($RVI$, $NDPI$, $DPSVIm$). Instead, in \cite{d2021parcel}, temporal features are derived solely from S1 data, resulting in a spatially and temporally consistent time series using both VV and VH polarizations, as well as the cross-polarization ratio (CR), which is calculated as the ratio of VH to VV backscatter. The backscatter values ($\sigma0$) are averaged over 10-day periods for each pixel. This is done separately for ascending and descending acquisitions and for both VV and VH polarizations. The averaged backscatter values are then converted to decibels (dB), which standardizes the data for further processing. The CR, which is the ratio of VH to VV backscatter, is calculated for each scene and then averaged over the same 10-day periods. The study creates a gridded time series of consistent temporal features for VV, VH, and CR, used as input data for machine learning models.
In contrast, the approach in the present work centers on spatial and textural feature extraction from S1 data, VH polarization. 
A set of spatial domain filters was applied to four seasonal composite images, also referred to as "super images." Each of these super images represents a specific season (winter, spring, summer, and autumn) and is generated by calculating the arithmetic mean of S1 images taken during that season. This approach allows for the aggregation of temporal information into a single representative image per season, enabling more efficient spatial analysis while preserving seasonal variability. The "super image" provides a single, characteristic view, onto which spatial filters are applied to highlight spatial patterns and textural features of the radar backscatter. This technique offers a spatial representation that enhances texture and spatial details, rather than focusing on temporal variability.
Moreover, regarding the calculation of the arithmetic mean, the number of S1 images varies from season to season, and this variability also depends on the area of interest. Due to the substantial volume of data in an annual time series over a large area of interest, a key advantage of the seasonal multitemporal approach is that there is no need to select a specific number of images based on strict spatial coverage criteria (e.g., 80$\%$, 90$\%$, or 100$\%$). By averaging the temporal data, even images with partial spatial coverage—such as those capturing only a corner or an edge—can be included in the analysis.\\
In summary, the linked studies mentioned above prioritize temporal consistency and temporal backscatter trends through multi-period averaging and cross-polarization, while the present work highlights spatial and textural details from a single seasonal composite, aiming at improving the spatial representation and at enhancing textural features in VH backscatter data.
By synthesizing features from available seasonal data, the used system adeptly captures the essential spatial characteristics necessary for accurate LC mapping, while effectively managing the variability introduced by less frequent imaging. Hence, the used network proves to be effective for global LC classifications, achieving robust results with a technique that efficiently summarizes spatial characteristics from sparser temporal data, tailored to suit the climatic characteristics of different regions given the 10m resolution employed.

In this work,  a novel approach combining a transformer-based Swin-Unet architecture with seasonal synthesized spatio-temporal images has been employed to classify LC types using spatio-temporal features extracted from S1 SAR data, organized into seasonal clusters. 
The workflow of the \textit{Swin-Unet} architecture is shown in Fig.~\ref{fig:swin_workflow}, and it was first introduced by \cite{cao2021swinunet} for medical image segmentation tasks. Unlike traditional CNNs, the \textit{Swin-Unet} architecture overcomes limitations in learning global and long-range semantic information by incorporating a patch partition layer and a Transformer-based U-shaped Encoder-Decoder architecture with skip connections. This allows for both local and global semantic feature learning. The encoder utilizes a hierarchical structure with a shifted window self-attention mechanism to extract context features, while a symmetric decoder with a patch-expanding layer performs the up-sampling operation, restoring the spatial resolution of the feature maps.
Recent advancements in computer vision, such as the integration of attention modules like CBAM in YOLOv8 architectures, have demonstrated the effectiveness of refining spatial features to enhance model performance in complex environments \cite{KHAN2024105195}. While these techniques are applied to object detection, our approach leverages the Swin-Unet model, which inherently incorporates hierarchical global attention through a shifted window mechanism, enabling precise spatio-temporal feature extraction for LC classification. This approach leads to superior results compared to those obtained using 2D and 3D CNN-based architectures like \textit{Attention U-Net} \cite{oktay2018attention} and 3D-FCN from \cite{GambaMarzi}. Another advantage is that the transformer model has a significantly lower number of trainable parameters (\textit{25 million}, compared to approximately \textit{31 million} and \textit{26.5 million}, respectively), leading to reduced computational costs, decreased memory requirements, and mitigated risk of overfitting.

Compared to other works, this study conducts LC mapping at a global level by analyzing three distinct regions of interest: Africa, Amazonia, and Siberia.


Unlike many studies that focus on specific and spatially limited regions, this work extends the analysis across diverse climatic zones and further refines the study areas into ecoregions, allowing for a detailed assessment of classification accuracy across different environmental conditions. Moreover, while existing approaches frequently rely on multisensor data (combining radar and optical sources) this study demonstrates that solely using S1 SAR data is sufficient to achieve remarkably high accuracy across all three test areas.

Another key distinction is that, whereas some methodologies depend on very high-resolution (VHR) radar data, this research effectively leverages S1 imagery at 10m resolution, yet still achieves optimal classification results. Furthermore, instead of employing dense, continuous temporal sequences as commonly done, this study introduces a novel feature organization strategy based on seasonal clustering. This innovative approach not only enhances land cover classification performance but also provides a robust solution for scenarios with limited temporal coverage, offering a more structured and integrative use of both spatial and seasonal temporal information.

Moreover, it is worth to highlight that this work was developed in the framework of  \href{https://climate.esa.int/en/projects/high-resolution-land-cover/}{ESA CCI+ HR
LC project, Phase 2}. This project focuses on understanding how LC and LC changes affect climate, aiming to improve climate modeling by examining the impact of spatial resolution on climate data. The LC analysis is crucial for measuring surface energy, water fluxes, greenhouse gas sources, and monitoring land use changes and extreme weather events. The project supports the Global Climate Observing System's Essential Climate Variables (ECVs) to aid the United Nations (UN) Framework Convention on Climate Change (UNFCCC). It aims to explore how varying spatial and temporal resolutions influence LC classification and to develop methods for generating and updating ECV products for long-term climate research.

The rest of this article is organized as follows. Section \ref{sec:methods} delves into the data pre-processing steps and neural network data preparation, detailing the extraction and computation of features of interest and the generation of the training set. Section \ref{sec:data_test_areas} examines the data sources and the three chosen macro areas for testing. Section \ref{sec:results} presents the outcomes across these selected study areas, including an assessment of performance within different ecoregion boundaries. Finally, Section \ref{sec:conclusions} discusses the advantages and potential future developments of this work.

\begin{figure*} [htp]
	\centering
	\adjustbox{max width=\textwidth}{\includegraphics[]{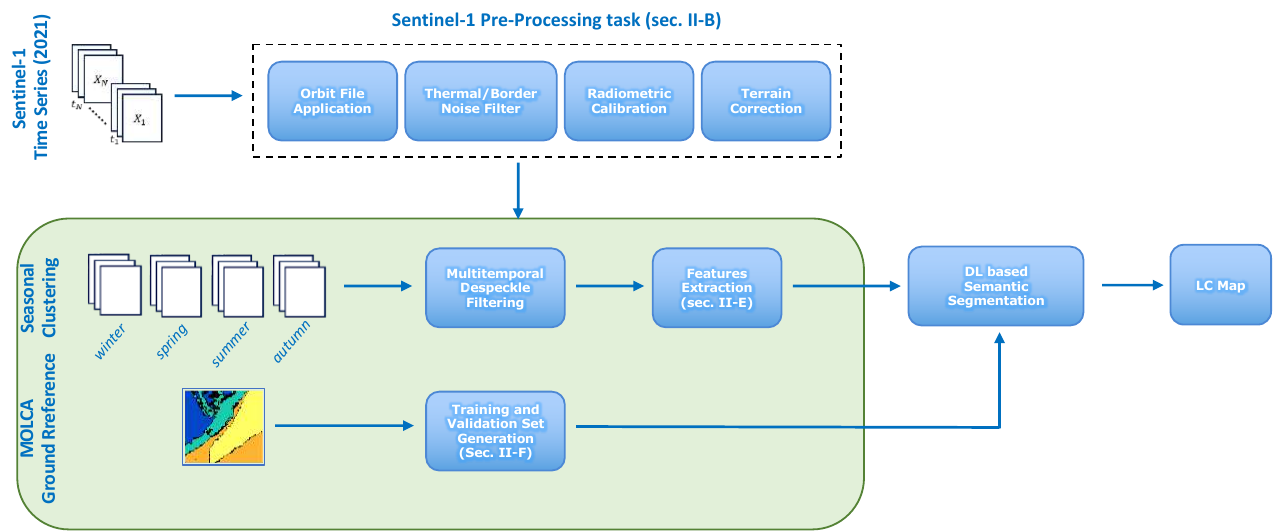}}
	\caption{A simplified workflow diagram of the proposed mapping procedure applied to SAR temporal sequences. The pre-processing part was done using a SNAP graph, as explained in Section \ref{subsec:data_preprocessing}. The multitemporal speckle noise reducer, feature extraction, training and validation set generation are also described in Section \ref{sec:methods}, while the DL-based block and the results are discussed in Section \ref{sec:results}.}
	\label{fig:workflow}
\end{figure*}

\subsection{Key Contributions}
To recap, the key contributions of this article include:

\begin{enumerate}
\item Development and utilization of synthesized spatio-temporal features derived from S1 SAR time series, organized into seasonal clusters. This approach effectively captures essential spatial and temporal characteristics without relying on dense temporal sequences, offering computational efficiency.
\item Enhanced LC classification performance through the use of seasonal clustering of weather-independent S1 SAR data, particularly in areas where optical data are significantly impacted by cloud coverage.
\item The innovative integration of the Swin-Unet model, which leverages hierarchical global attention to improve feature extraction and spatial representation.
\item Application of a global-level HR LC mapping approach across diverse regions (e.g., Africa, Amazonia, Siberia) and their respective ecoregions. This method achieves high classification accuracy using 10m resolution S1 SAR data, even in regions with limited training data.
\item Extensive validation across distinct ecoregions, showcasing the model’s generalization capability and robust classification accuracy in diverse environmental contexts.
\item Advancement of research within the framework of the ESA CCI+ HR LC project, Phase 2. The work emphasizes the significance of LC changes for climate studies and contributes to the generation of Essential Climate Variables (ECVs) for long-term climate monitoring.
\end{enumerate}

\section{The proposed approach}
\label{sec:methods}

The proposed approach is based on SAR S1 sequence classification using a DL architecture. Specifically, a SAR sequence is subdivided into seasonal subsequences, and a set of spatial features are computed from these reduced time series. To these spatio-temporal features the aforementioned \textit{Swin-Unet} is then applied. The overall b lock diagram of the described procedure is shown in Fig. \ref{fig:workflow}.

The proposed methodology is applied to spatial subsets (tiles) according to the tiling system used by ESA for S2 \cite{gatti2018sentinel}. This allows for a uniform coverage of geographically large areas, and potentially the entire Earth's surface. Indeed, in this paper the approach is applied to S1 SAR datasets in test areas covering Amazonia, Africa, and Siberia. These three areas were selected to assess the robustness of the methodology in regions with significantly different environmental and climatic characteristics, as discussed in more detailes in Section \ref{subsec:areas}.

\subsection{SAR Sentinel-1 data pre-processing}
\label{subsec:data_preprocessing}

\begin{figure} [htp]
    \centering
    \includegraphics[width=.5\columnwidth]{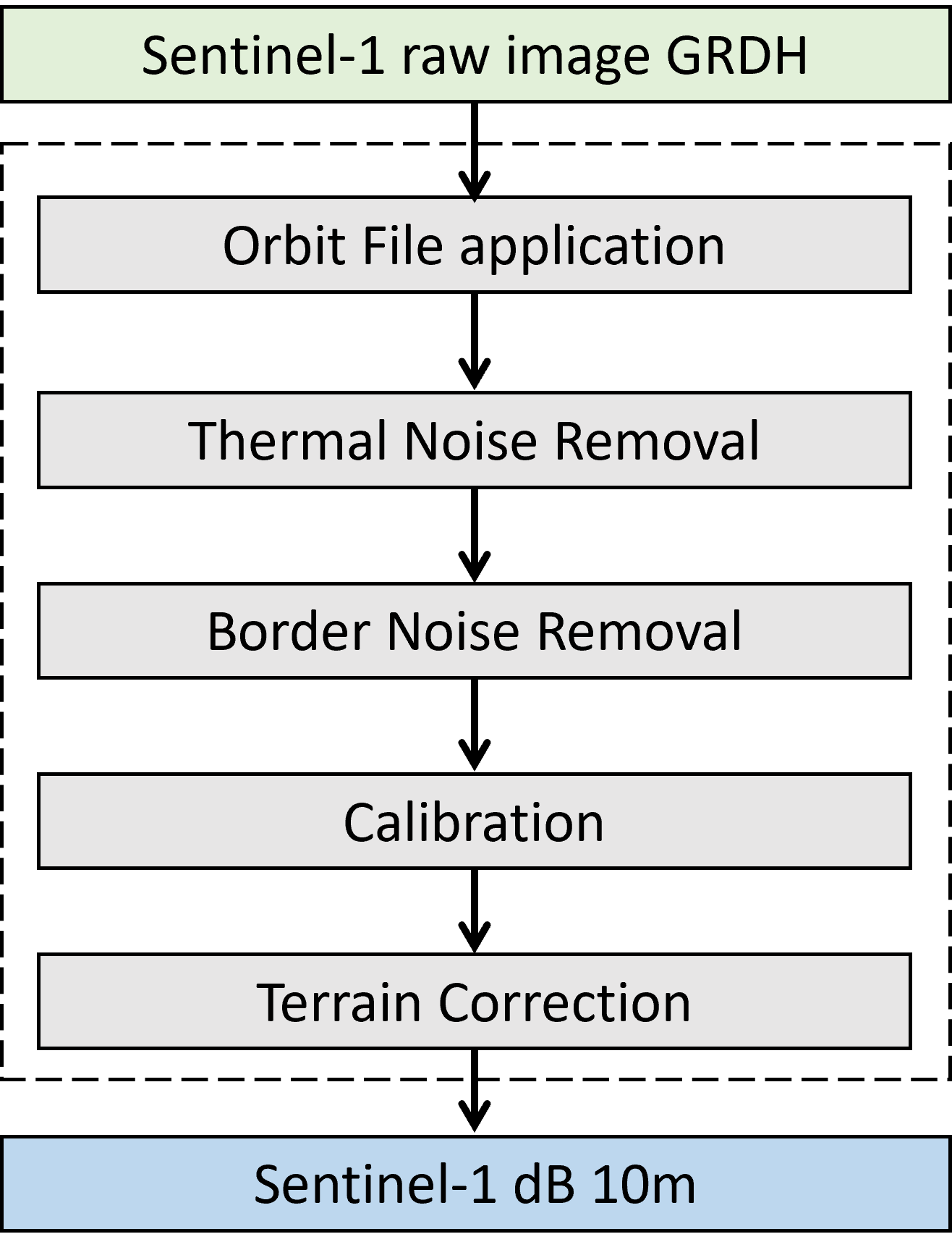}
    \caption{Block diagram of S1 data pre-processing.}
    \label{fig:sar_preprocessing}
\end{figure}

The initial step involves pre-processing the radar sequences, similar to the method employed by the European Space Agency's (ESA) Sentinel Application Platform. \cite{foumelis2018esa}. Each considered S1 sequence is assumed as composed by multiple images taken on the same orbit with the same beam and polarization. These images are preprocessed according to a set of correction and refinement steps \cite{sorriso2021general} and shown in Fig.~\ref{fig:sar_preprocessing}.

The chain in Fig.~\ref{fig:sar_preprocessing}, consists of a number of steps applied to the Level-1 S1 data, as follows:

\begin{itemize}
    \item \textbf{Orbit File application}: to correct the satellite position and add velocity information, in order to achieving a  precisely geocoded observations.
	\item \textbf{Thermal Noise removal}: thermal noise is an additive interference component superimposed on the signal of interest and is processed with the same processing gains applied to the true signal. Thermal noise removal reduces the effects of the thermal distortion in the inter-sub-swath texture and performs the normalization of the backscattered signal within the entire acquisition.
	\item \textbf{Border Noise removal}: the Border Noise Removal algorithm \cite{filipponi2019sentinel} removes the low intensity noise caused by radiometric artefacts and invalid data at the edges of the image.
	\item \textbf{Radiometric calibration}: radiometric calibration is necessary because the grey level of SAR imagery must be adjusted taking into account the backscatter signals from objects in the area. The digital number of the pixel is converted to a radiometrically calibrated backscatter value.
	\item \textbf{Geometric Terrain correction}: terrain correction eliminates the distortion caused by the topographical variations \cite{699700}, thereby increasing the accuracy of the location of the objects in the scene. The correction is performed by using the Digital Elevation Model (DEM) data to extract the height information of the object. The data is then resampled using nearest-neighbour interpolation at a pixel spacing of 10 m spatial resolution and geolocated in the WGS84 coordinate system.

\end{itemize}

\begin{figure} [htp]
    \centering
    \includegraphics[trim={1cm .5cm 2cm 1cm},width=8cm]{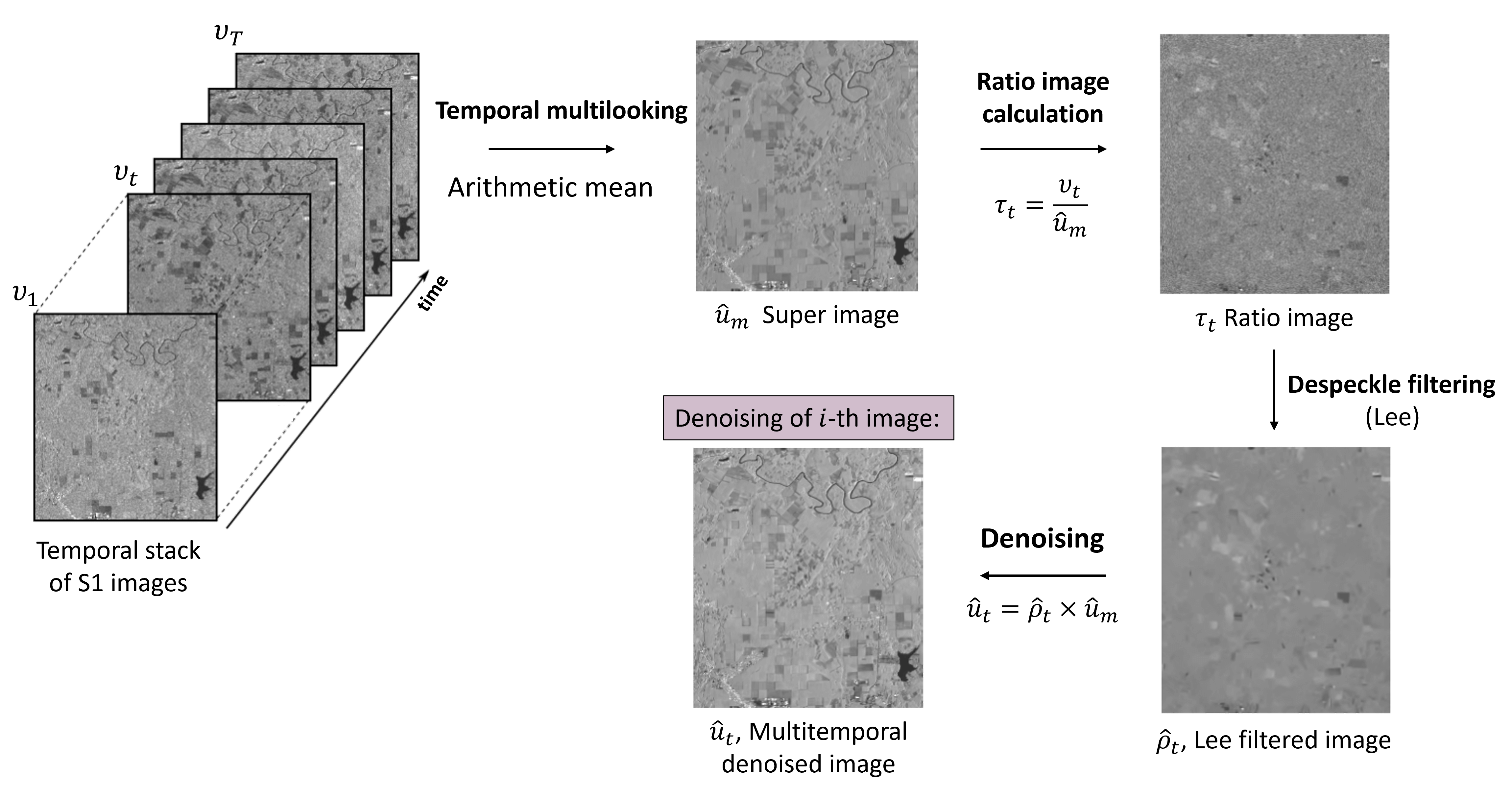}
    \caption{Multitemporal despeckle flowchart applied to the S1 temporal sequence. The temporal averaging of  the SAR time series produces the super image $\hat{u}_m$. The super image is used to form the ratio image $\tau_t$, given by the image $\upsilon_t$ at time $t$ and the super image $\hat{u}_m$, for each pixel of the S1 images. The Lee filter is then applied to $\tau_t$ because the super image $\hat{u}_m$ suffers from speckle (although the speckle in the super image is greatly reduced), resulting in the image $\hat{\rho}_m$. In the last step, the restored image $\hat{u}_t$ is obtained by multiplying the denoised ratio image with the super image.}
    \label{fig:multi_denoising}
\end{figure}

\subsection{Feature extraction}
\label{sec:features}

Once the sequence has been preprocessed and somehow ``harmonized", spatio-temporal SAR features are extracted and used as inputs to the DL architecture. Specifically, following \cite{marzi2020global}, it is assumed that temporally aggregated portions of a full annual time series are more appropriate for LC mapping. In this approach the original SAR sequence is subdivided into four ``seasonal'' clusters that approximate the seasonal cycle of the different LC types. These identified clusters suffer from speckle distortion due to the nature of SAR acquisition \cite{oliver2004understanding}, which affects the appearance of the image and the performance of the scene analysis. Speckle in SAR is a multiplicative effect (i.e. it is directly proportional to the grey value of the pixel) and is mitigated using a multitemporal denoising filter, as described below, to preserve radiometric and textural information.

\subsubsection{Multitemporal Despeckle filtering}
\label{subsec:despeckle_filtering}

A suitable and advanced multitemporal denoising filter based on the one described in \cite{zhao2019ratio} is applied to the four ``seasonal'' clusters. The multitemporal approach appears to provide better results than a spatial filter applied independently to each SAR image, thanks to the exploitation of the temporal sequence, to the benefit of a better spatial resolution preservation. The filter is ratio-based and computes an image, called \textit{super image}, by exploiting the SAR time series. In fact, the temporal averaging of SAR time series produces the super image $\hat{\mathbf{u}}_m$, according to the formula below:
\begin{equation*}
    \centering
    \hat{\mathbf{u}}_m=\frac{1}{S}\sum_{s=1}^{S}{\mathbf{x}_s} \ ,
\end{equation*}
    where $S$ is the images's number of the seasonal sequence, $\mathbf{x}_s$ is the SAR image of the seasonal sequence, and $s$ is the time index.

Speckle is reduced and spatial resolution is preserved. The filtered image is thus recovered by exploiting the statistical properties associated with the original super image.
Basically, the method comprises three steps: a) computation of the seasonal super image $\hat{\mathbf{u}}_m$ by the arithmetic averaging of time series images; b) denoising of the ratio image; c) calculation of the final images given by the multiplication between the denoised ratio and the super image. These steps and more details are well reported in the workflow shown in Fig.~\ref{fig:multi_denoising}.

\subsubsection{Spatio-temporal feature computation}
\label{subsec:features_extraction}

After the data pre-processing described in Section~\ref{subsec:data_preprocessing}, a spatio-temporal feature extraction is performed using the polarimetric information derived from the intensities of the VH (cross-polarized) S1 polarization. Rather than considering complex spatial features such as shape and size, which would require unsupervised segmentation of the image, a set of textural feature set is computed computed for each of the four seasonal super image $\hat{\mathbf{u}}_m$:

\begin{itemize}
    \item Lee filter, an adaptive filter recognized as the first model-based approach specifically designed for reducing speckle noise \cite{sorriso2021general}. It maintains significant edges, linear structures, point targets, and texture details by minimizing the mean square error.
    \begin{equation*}
        \centering
        u_{\mathrm{LEE}}(i,j)=\bar{u}_m(i,j)+K \cdot (\hat{u}_m(i,j)-\bar{u}_m) \ ,
    \end{equation*}
    where $u_\mathrm{{LEE}}(i,j)$ is the Lee-filtered pixel value at position $(i,j)$, $\bar{u}_m$ is the local mean of the pixels within a specified kernel $K$ around the pixel $(i,j)$, $\hat{u}_m(i,j)$ is the seasonal super image at location $(i,j)$, and $K$ is the weighted factor given by the noise variance and the local variance within the window.\\
    \item Median filter, is a non-adaptive filter that substitutes each pixel's value with the median of the values from the surrounding local neighborhood:
    \begin{equation*}
        \centering
        u_{\mathrm{MEDIAN}}(i,j)=\mathrm{median} \{\hat{u}_m(i',j') \ | \ (i',j') \in\mathcal{N}(i,j)\} \ ,
    \end{equation*}
    where $\mathcal{N}(i,j)$ is the neighborhood defined by the kernel centered at pixel $(i,j)$, the set $\{\hat{u}_m(i',j')\}$ contains the pixel values within the kernel surrounding the pixel $(i,j)$. The median function selects the middle value from the sorted pixel values in the neighborhood.\\
    \item Mean filter, is among the most commonly utilized low-pass filters (LPF), replacing the value of the pixel with the average of all the values within the surrounding local neighborhood (filter kernel):
    \begin{equation*}
        \centering
        u_{\mathrm{MEAN}}(i,j)=\frac{1}{N}\sum_{(i',k')\in\mathcal{N}}\hat{u}_m(i',k') \ ,
    \end{equation*}
    where $u_{\mathrm{MEAN}}(i,j)$ is the Mean-filtered image at pixel $(i,j)$, $\mathcal{N}$ is the neighborhood defined by the kernel size, $N$ is the total number of pixels in the neighborhood $\mathcal{N}$, and $(i',k')$ are the coordinates of the pixels in the neighborhood around $(i,k)$.\\
    \item Maximum (or Minimum) filter, non-linear filter identifies the brightest (or darkest, for the minimum) point in an image. It is based on the median filter, as it corresponds to the $100th$ (or $0th$, for the minimum) percentile, meaning it selects the maximum (or minimum) value of all the pixels within a specified local region of the image.
    \begin{align*}
        \centering
        u_{\mathrm{MAX}}(i,j)=\max_{(i',j')\in\mathcal{N}} \hat{u}_m(i',j') \ & , \\
        u_{\mathrm{MIN}}(i,j)=\min_{(i',j')\in\mathcal{N}} \hat{u}_m(i',j') \ & .
    \end{align*}
    \item Range (Max-Min) filter, enhances image contrast by calculating the difference between the dilation and erosion (maximum and minimum) of the original image. For the S1 seasonal super image $\hat{\mathbf{u}}_m$, the filtered output $\mathbf{u_{\mathrm{Max-Min}}}$ is given by:
    \begin{equation*}
        \centering
        \mathbf{u_{\mathrm{MAX-MIN}}}=\mathbf{u_{\mathrm{MAX}}}-\mathbf{u_{\mathrm{MIN}}} \ ,
    \end{equation*}
    where $\mathbf{u_{\mathrm{MAX}}}$ and $\mathbf{u_{\mathrm{MIN}}}$ are the output of applying maximum (dilation) and minimum (erosion) filters to the input super image $\hat{\mathbf{u}}_m$, respectively, as defined above.
\end{itemize}

A kernel of $5 \times 5$ pixels is used to derive each of these spatial statistical descriptors. 

In summary, after the SAR sequence is properly pre-processed, four subsequences are selected and the corresponsing super images super are computed. Finally, for each superimage seven spatial features are extracted, resulting in 28 features to be used for the classification.

\subsection{Network pre-processing}
\label{proposed_approach}

Before being fed into the DL archirecture, to the aforementioned features a \textit{min-max normalization} is applied, to ensure that all data falls within the same scale range. This is performed because unscaled input variables can result in a slow or unstable learning process \cite{bishop1995neural}.

Additionally, \textit{data augmentation} techniques are employed since, as expressed in \cite{10399888}, DL architectures (especially ViTs)  require large training datasets to maximize their classification accuracy. Specifically, data augmentation is implemented by dividing both the input and reference images into smaller patches, each measuring 256 pixels, with an overlap determined by a stride of 128 pixels. This approach enhances dataset diversity and increases the number of training samples. By generating smaller patches, the method improves dataset coverage, which is particularly beneficial when only a limited number of tiles are available, such as in the context of ecoregion analysis (see Section \ref{ecoregion_section}).

Increasing the dataset size also helps in preventing overfitting issues and in reducing memory usage. Utilizing smaller patches instead of processing entire tiles with the original size (512x512 pixels) minimizes memory requirements and computational burden, thereby improving the efficiency of the training process.

Following the aforementioned patching process, the number of samples is increased significantly across all regions. In Siberia, the initial dataset of $64$ tiles is expanded to $576$ patches. Similarly, Africa grows from $103$ tiles to $927$ patches, while Amazonia increases from $86$ tiles to $774$ patches. This increase in sample availability strengthens model training by providing a more comprehensive representation of spatial patterns, improving generalization, and enhancing robustness across different regions.

\begin{figure*} [htp]
	\centering
	\adjustbox{max width=.6\textwidth}{\includegraphics[]{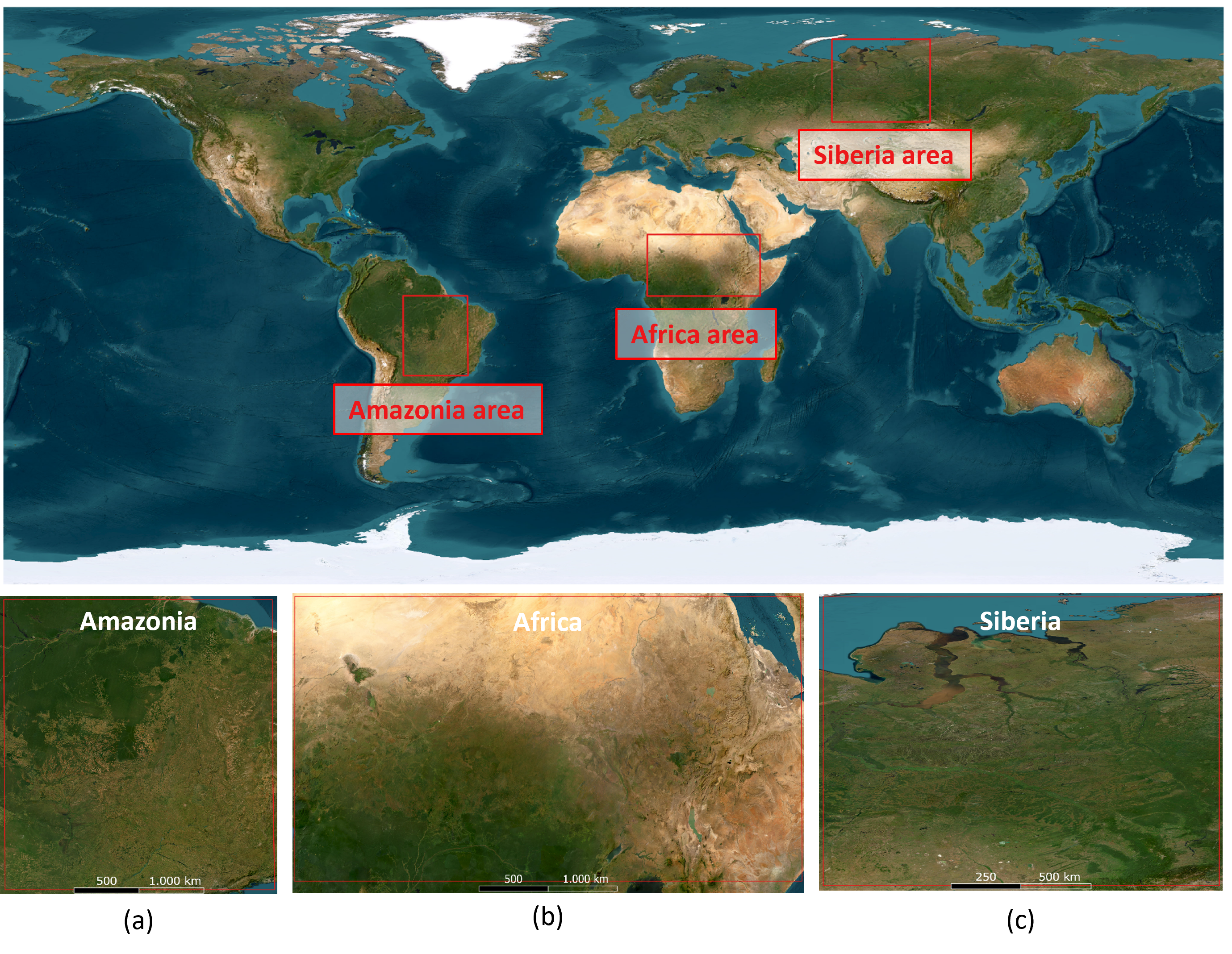}}
	\caption{Enlargements of the test areas: (a) Amazonia (62.1014$^\circ$ W, 23.5983$^\circ$ S : 42.9441$^\circ$ W, 0$^\circ$ N, WGS 84), (b) Africa (9.8986$^\circ$ E, 0.0885$^\circ$ S : 43.2908$^\circ$ E, 18.0891$^\circ$ N, WGS 84) and (c) Siberia (64.4361$^\circ$ E, 51.2789$^\circ$ N : 93.4017$^\circ$ E, 75.6847$^\circ$ N, WGS 84).}
	\label{fig:test_areas}
\end{figure*}

\subsection{Swin-Unet} \label{sec:swin_unet}
As introduced in Section \ref{intro}, the best performing DL model was used for the classification task at hand is the \textit{Swin-Unet} \cite{cao2021swinunet}. This architecture consists of three main blocks: the \textit{encoder}, the \textit{bottleneck}, and the \textit{decoder}.

The \textit{encoder} utilizes multiple \textit{Swin Transformer} layers \cite{liu2021swin}, designed to process the input images hierarchically through a series of stages. Each stage implements the \textit{shifted window self-attention} \cite{yu2022selfattention}. This mechanism allows the model to efficiently capture local interactions in the early stages and progressively build up to understanding broader areas of the input image. This effectively reduces the resolution of feature maps while increasing the feature dimensions, enabling a deep and rich understanding of the input data.

The \textit{bottleneck} serves as the critical transition point between the encoder and decoder modules. It is typically composed of one or more \textit{Swin Transformer} layers situated at the deepest part of the network. This middle part of the network focuses on integrating and compressing the high-level features learned by the encoder.

Finally, the \textit{decoder} focuses on progressively expanding the encoded features back to the original image resolution. It employs \textit{Swin Transformer layers} arranged in stages, each one using a patch expanding layer to progressively increase the spatial resolution of the feature maps. Additionally, \textit{skip connections} from corresponding encoder stages are integrated at each level of the decoder. These connections help restore spatial details that are often lost during down-sampling in the encoder. A simple diagram of the overall model is shown in Fig. \ref{fig:swin_workflow}.

\subsection{Swin-Unet model configuration} \label{sec:swin_unet_cfg}
The Swin-Unet model configuration was carefully designed to balance computational efficiency and model performance. Among the selected parameters are \textit{feature size}, \textit{batch size}, and \textit{learning rate}.

The \textit{feature size} parameter determines the dimensionality of the initial feature maps and directly impacts the model's complexity. For this study, the value was set to $48$, resulting in approximately $25$ million trainable parameters, as shown in Table \ref{table:model_fps_comparison} under the \textit{Complexity} column. A more detailed discussion of model complexity, including additional parameters influencing computational demands, is presented in Section \ref{sec:discussions}. This choice represents a balance between model capacity and computational overhead. Although higher values such as $60$ or $72$ could enhance the model's representational capability, they also introduce significantly higher memory requirements and computational costs. Testing with values greater than $48$, such as $60$ and $72$, did not result in significant accuracy improvements, but exponentially increased the model's complexity, reaching approximately $39.2$ and $56.5$, respectively.

The \textit{batch size} for model training was fixed at $1$ due to the high-dimensional nature of the input data, which consists of $28$-channel spatio-temporal features. This choice was influenced by memory constraints, as larger batch sizes would exceed the available GPU memory.

The \textit{learning rate} was set to \textbf{$10^{-4}$}, ensuring stable and efficient optimization, achieving consistent improvements on the validation set. Training was carried out for a total of 30 \textit{epochs}, following the parameter hyperparameter configuration detailed in \cite{GambaMarzi}. Specifically, the model utilized the \textit{Adam} optimizer and \textit{Categorical Cross-Entropy} loss function for LC classification optimization. These settings allowed the model to handle the 28-feature temporal sequence efficiently and ensured stable performance during training.

By setting \textit{feature size} to $48$, \textit{batch size} to $1$, and \textit{learning rate} to $10^{-4}$, the Swin-Unet model achieved an optimal balance between computational feasibility and predictive accuracy. The total complexity of the model, measured as the number of trainable parameters, ensured the capacity to efficiently handle the LC classification task while maintaining computational efficiency. 
This configuration was validated through experimental runs across various dataset separations, as shown in Table \ref{results_random_state}, demonstrating stable convergence and robust accuracy on the test set, regardless of the data split.

\section {Input data and test areas} \label{sec:data_test_areas}

As mentioned earlier, in this work the general procedure described in the previous section has been applied to S1 datasets in specific geographical locations. 

\subsection{Sentinel-1 mission and data}
\label{sec:S1}

The S1 mission is the radar component of the European Copernicus programme, which has many operational applications in addition to the LC mapping. It is composed by a constellation of two satellites, S1A and -1B, with a C-band Synthetic Aperture Radar sensor. Each satellite is equipped with right-looking antennas with an angle of incidence between 29.1$^\circ$ and 46$^\circ$ and has a revisit time of approximately 6 days over land (equatorial in a polar orbit).


The data are also available free of charge to the public via the Copernicus Open Access Hub at \url{https://browser.dataspace.copernicus.eu/}, making it more attractive for new challenging applications and opportunities \cite{hu2018feature}.

For this work, level-1 Ground Range Detected (GRD) products acquired in Interferometric Wide Swath (IW) mode were used. In these products the phase information is lost due the application of the multi-looking filter and the ground range projection based on an Earth ellipsoid model. The datasets are in HR and provide images with a native range by azimuth resolution  $20 \times 22$ m and pixel resolution equals to $10 \times 10$ m.\\
In this study, the VH polarization has been selected, as outlined in \cite{marzi2023automatic} and \cite{sorriso2021general}, due to its heightened sensitivity to vegetation, water and urban characteristics \cite{ruetschi2017using} \cite{marzi2021inland} \cite{marzi2024joint}. Furthermore, to strengthen this choice, tests were conducted using the VH polarization alone, as well as in combination with the VV polarization. The comparative results of these tests are documented in the Product Validation and Algorithm Selection Report (PVASR) for the \href{https://climate.esa.int/en/projects/high-resolution-land-cover/}{ESA CCI+ HR
LC project, Phase 1}, which is freely accessible on the dedicated website under the '\textit{Key Documents}' section. The findings indicated that the inclusion of the VV polarization did not provide a consistent improvement in LC classification, thereby reinforcing the preference for the VH band.

\begin{figure*}[!ht]
	\centering
	\includegraphics[width=2\columnwidth]{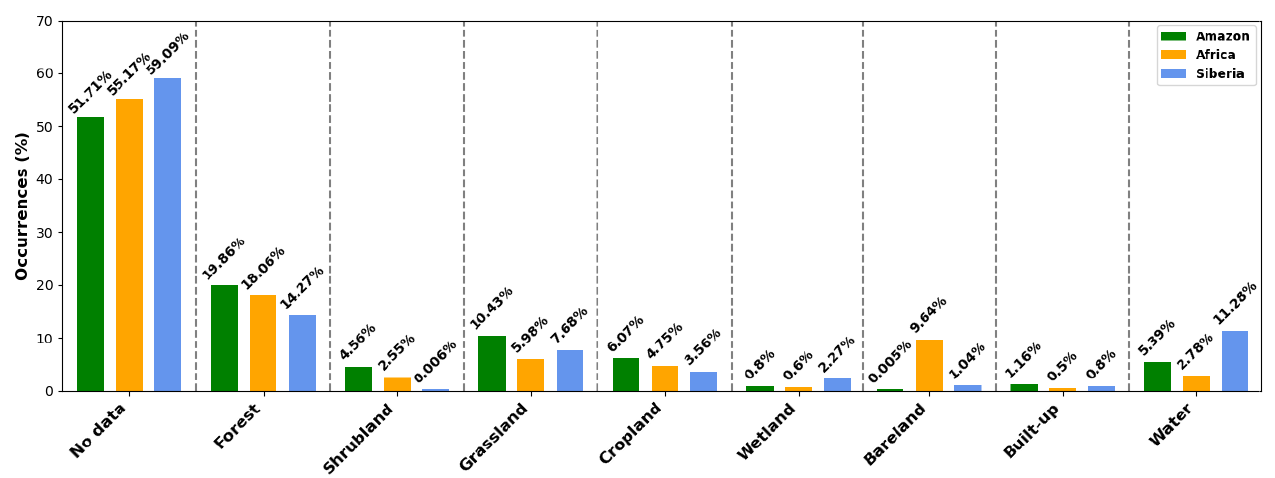}
	\caption{Distribution of classes for the three areas of interest.}
	\label{class_distrib}
\end{figure*} 

\begin{figure*} [htp]
        \setlength{\abovecaptionskip}{-10pt}
	\begin{minipage}[t]{\textwidth}
		\centering
		\includegraphics[width=\linewidth]{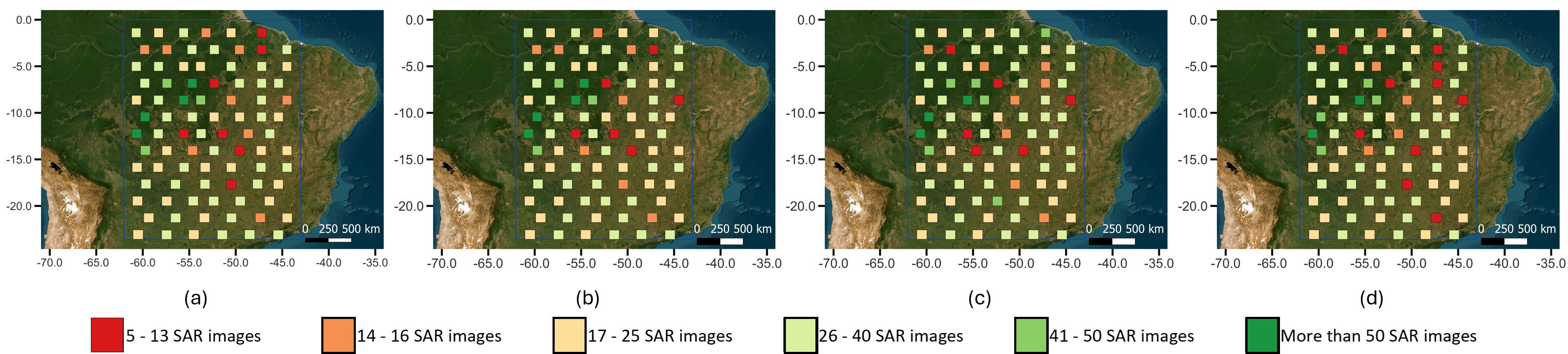}
		\caption{Seasonal distribution of S1 acquisitions in 2021 for the S2 tiles selected to collect the training set over the Amazonia site. The distribution is shown for: (a) winter, (b) spring, (c) summer and (d) autumn seasons.}
        \vspace{25pt}
		\label{figure:Amazonia_s1_distribution}
	\end{minipage}
        \setlength{\abovecaptionskip}{-10pt}
        \begin{minipage}[t]{\textwidth}
		\centering
		\includegraphics[width=\linewidth]{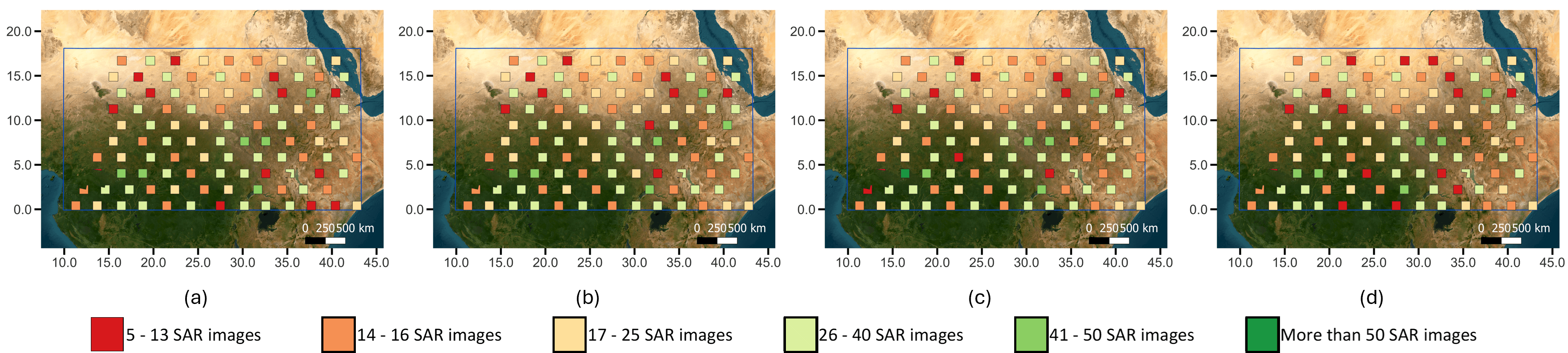}
		\caption{Seasonal distribution of S1 acquisitions in 2021 for the S2 tiles selected to collect the training set over the Africa site. The distribution is shown for: (a) winter, (b) spring, (c) summer and (d) autumn seasons.}
        \vspace{25pt}
		\label{figure:africa_s1_distribution}
	\end{minipage}
        \setlength{\abovecaptionskip}{-10pt}
	\begin{minipage}[t]{\textwidth}
		\centering
		\includegraphics[width=\linewidth]{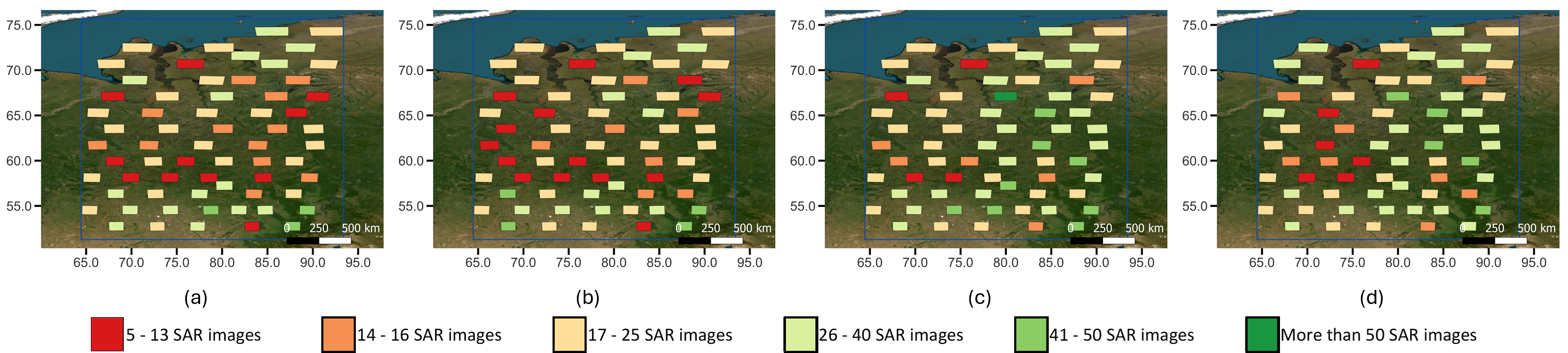}
		\caption{Seasonal distribution of S1 acquisitions in 2021 for the S2 tiles selected to collect the training set over the Siberia site. The distribution is shown for: (a) winter, (b) spring, (c) summer and (d) autumn seasons.}
		\label{figure:siberia_s1_distribution}
        \end{minipage}
\end{figure*}

\subsection{Test areas}
\label{subsec:areas}
Three test macro areas in Amazonia, Africa, and Siberia (see Fig.~\ref{fig:test_areas}) have been identified and used for the experiments, according to the guidelines of the European Space Agency (ESA) ``Climate Change Initiative Extension (CCI+) Phase 2: New Essential Climate Variables (NEW ECVS)'' project. 

These areas correspond to very different LC and climate typologies, and the selected portions have a wide extension each (5.370 billion km$^2$ in Amazonia, 7.340  billion km$^2$ in  Africa, and 3.787 billion km$^2$ in  Siberia).

\begin{table*}[htp]
	\centering
	\caption{Class legend of MOLCA dataset with the number of samples per region (train and test set)*.}
	\begin{tabular}{lccccccccccccccccccc}
		\toprule
		\textbf{MOLCA label} & \textbf{LC type} & \textbf{Color} & \multicolumn{2}{c}{\textcolor{black}{\textbf{Africa}}} & \multicolumn{2}{c}{\textcolor{black}{\textbf{Siberia}}} & \multicolumn{2}{c}{\textcolor{black}{\textbf{Amazonia}}} \\
		&  &  & \textcolor{black}{\textbf{Train}} & \textcolor{black}{\textbf{Test}} & \textcolor{black}{\textbf{Train}} & \textcolor{black}{\textbf{Test}} & \textcolor{black}{\textbf{Train}} & \textcolor{black}{\textbf{Test}} \\
		\midrule
		20 & Forest & \cellcolor[HTML]{006400} & \textcolor{black}{7,845,129} & \textcolor{black}{3,171,303} & \textcolor{black}{3,734,223} & \textcolor{black}{1,721,781} & \textcolor{black}{6,780,161} & \textcolor{black}{3,168,114} \\
		5 & Shrubland & \cellcolor[HTML]{966400} & \textcolor{black}{1,135,099} & \textcolor{black}{479,830} & \textcolor{black}{958} & \textcolor{black}{539} & \textcolor{black}{1,689,319} & \textcolor{black}{627,496} \\
		7 & Grassland & \cellcolor[HTML]{ffb432} & \textcolor{black}{2,486,506} & \textcolor{black}{1,143,817} & \textcolor{black}{1,909,583} & \textcolor{black}{980,305} & \textcolor{black}{3,911,300} & \textcolor{black}{1,349,862} \\
		8 & Cropland & \cellcolor[HTML]{ffff64} & \textcolor{black}{1,961,002} & \textcolor{black}{904,454} & \textcolor{black}{946,878} & \textcolor{black}{357,745} & \textcolor{black}{1,967,779} & \textcolor{black}{973,989} \\
		9 & Wetland & \cellcolor[HTML]{1bcbae} & \textcolor{black}{281,537} & \textcolor{black}{103,895} & \textcolor{black}{657,152} & \textcolor{black}{243,270} & \textcolor{black}{167,227} & \textcolor{black}{291,961} \\
		11 & Lichens and mosses & \cellcolor[HTML]{ffdcd2} & \textcolor{black}{0} & \textcolor{black}{0} & \textcolor{black}{0} & \textcolor{black}{0} & \textcolor{black}{0} & \textcolor{black}{0} \\
		12 & Bareland & \cellcolor[HTML]{9b969b} & \textcolor{black}{4,129,099} & \textcolor{black}{1,756,976} & \textcolor{black}{321,283} & \textcolor{black}{106,774} & \textcolor{black}{2,365} & \textcolor{black}{879} \\
		13 & Built-up & \cellcolor[HTML]{c31400} & \textcolor{black}{177,943} & \textcolor{black}{97,767} & \textcolor{black}{211,478} & \textcolor{black}{100,329} & \textcolor{black}{445,613} & \textcolor{black}{171,281} \\
		15 & Water & \cellcolor[HTML]{0046c8} & \textcolor{black}{1,097,124} & \textcolor{black}{457,770} & \textcolor{black}{2,831,258} & \textcolor{black}{1,463,209} & \textcolor{black}{2,139,339} & \textcolor{black}{644,332} \\
		16 & Permanent ice and snow & \cellcolor[HTML]{ffffff} & \textcolor{black}{0} & \textcolor{black}{0} & \textcolor{black}{0} & \textcolor{black}{0} & \textcolor{black}{0} & \textcolor{black}{0} \\
		\bottomrule
	\end{tabular}
	\label{table:MOLCA_legend}
	\vspace{5mm}
	\begin{minipage}{\textwidth}
		\centering
		\tiny{\textit{*Note: the numbers represent the number of pixel samples used for training and testing in each region.}}
	\end{minipage}
\end{table*}

\begin{itemize}
    \item Amazonia is a region dominated by vegetation and hot tropical weather, a perfect example of the utility of SAR data: indeed, it is very difficult to obtain cloud-free optical images over this area due to the harsh climate, with precipitation ranging from 200 to 320 mm per month and an average humidity of 89\%.
    \item Africa is a very complex, climate-sensitive region with a history of severe weather events, many of which are linked to global warming. Morphologically, the region is characterised by bare soil, lakes and arable land, and many zones have experienced severe droughts worth a deeper investigation.
    \item Finally, Siberia has a very cold climate year round, making it a potential hotspot for future climate change research. The region is characterized by many rivers and water bodies covered with ice and snow for about 75\% of the year. Here, the advantage of multitemporal SAR is clear: SAR signals can penetrate clouds and rain, ensuring periodic data acquisitions.
\end{itemize}

\subsection{Training set generation}
\label{subsection:molca_set}

To build the training set for the DL architecture in these areas, the Map Of LC Agreement (MOLCA) \cite{gorica_bratic_2023_8071675} has been used. MOLCA was generated using already existing global HR LC maps, retaining only those areas where all datasets agree on the same LC class and discarding areas of disagreement (these pixels are identified as nodata and are set to zero on the map). The MOLCA images, arranged according to the S2 Level-1C product tiling grid and distributed in GTiff format, cover the above mentioned 3 regions in Amazonia, Africa and Siberia, with about 117 billion pixels at 10m resolution.

The MOLCA dataset was produced as part of the European Space Agency (ESA) - funded Climate Change Initiative Extension (CCI+) Phase 1 New Essential Climate Variables (NEW ECVS) HR LC ECV (HR\_LandCover\_cci) project, known as CCI HRLC or CCI+ HRLC (\url{https://climate.esa.int/en/projects/high-resolution-land-cover/}).\\ The LC classes represented in MOLCA are shown in Table \ref{table:MOLCA_legend}, and cover the period from 2016 to 2020. The Table \ref{table:MOLCA_legend} also includes the number of samples (pixels) used for the train and test sets, which were employed in subsequent evaluations with DL models for each analyzed region: Africa, Amazonia, and Siberia. Notably, no representative samples (pixels) for the classes "Permanent Ice and Snow" and "Lichens and Mosses" are present in the three areas of interest.
The accuracy estimate for MOLCA indicates an Overall Accuracy (O.A.) of 96\% \cite{gorica_bratic_2023_8071675}.

As highlighted in Fig. \ref{class_distrib}, \textit{No data} values accounts for over 50$\%$ of the entire dataset in the study areas. For this reason, prior to the training phase, all the pixels in the input SAR sequences that correspond to this class should be set to zero. This approach helps prevent the model from learning erroneous relationships for the \textit{No data} class, which is useless and misleading.

Subsequently, to select a significant training data set, the areas in Fig. \ref{fig:test_areas} were randomly and homogeneously sampled in accordance with the S2 tiling, the spatial coverage of which is shown in Figures \ref{figure:Amazonia_s1_distribution}, \ref{figure:africa_s1_distribution}, and \ref{figure:siberia_s1_distribution} for Amazonia, Africa, and Siberia, respectively. Each tile with a size of $10980 \times 10980$ pixels in the UTM coordinate reference system has been patched in smaller areas of $549 \times 549$ pixels, i.e. 1/20th of the tile linear dimensions. The most significant patch, i.e., with the largest number of LC classes is then selected by means of visual inspection for each tile and area in order to obtain a balanced representation of the LC classes present in the scenes, the distribution of which is shown in Fig. \ref{class_distrib}.

Once the most representative patches have been identified, the corresponding S1 features are computed according to the methodology presented in the previous Section. The seasonal spatial distributions with respect to the availability of the S1 acquisitions are shown in Figures \ref{figure:Amazonia_s1_distribution}, \ref{figure:africa_s1_distribution} and \ref{figure:siberia_s1_distribution}, for Amazonia, Africa and Siberia, respectively. In the graphs, the adopted colormap represents varying levels of data availability, ranging from areas with only 5–13 images (indicated in red) to regions with more than 50 acquisitions (represented in dark green). The gradient between these colors highlights intermediate values, allowing for a visual understanding of data distribution. Additionally, the macro areas of interest are marked with blue rectangles.
Despite the presence of red tiles in each season, the number of acquisitions is sufficient to carry out the spatio-temporal feature extraction \cite{sorriso2021general}. The final training sets consist of 86 MOLCA patches and 2408 S1 features for the Amazonian area; 103 MOLCA patches and 2884 S1 features for Africa and 64 MOLCA patches and 1792 S1 features for Siberia.

\begin{table*}[htp]
	\centering
	\caption{Overall accuracy (O.A.), Kappa Coefficient, F1-Score and Producer Accuracy (PA) for the considered models in different regions.}
	\begin{tabular}{lccccccccccccc}
		\toprule
		\textbf{Region} & \textbf{Model} & \textbf{O.A.} & \textbf{kappa} & \textcolor{black}{\textbf{F1-Score}} & $\textbf{pa}_0$ & $\textbf{pa}_1$ & $\textbf{pa}_2$ & $\textbf{pa}_3$ & $\textbf{pa}_4$ & $\textbf{pa}_5$ & $\textbf{pa}_6$ & $\textbf{pa}_7$ & $\textbf{pa}_8$\\
		\midrule
		\multirow{4}{*}{Amazonia} 
        & Random Forest & 0.622 & 0.328 & 0.565 & 0.9 & 0.51 & 0.09 & 0.25 & 0.07 & 0.18 & 0 & 0.02 & 0.18 \\
        & 3D-FCN & 0.818 & 0.709 & 0.795 & 0.95 & 0.89 & 0.08 & 0.49 & 0.74 & 0.24 & 0 & 0.06 & 0.75\\
        & Attention U-Net & 0.843 & 0.760 & 0.832 & 1 & 0.91 & 0 & 0.61 & 0.6 & 0  & 0 & 0 & 0.96 \\
		& Swin-Unet & \textbf{0.933} & 0.898 & 0.924 & 1 & 0.97 & 0.5 & 0.88 & 0.85 & 0.03 & 0 & 0.85 & 0.99  \\
		\midrule
		\multirow{4}{*}{Africa} 
        & Random Forest & 0.745 & 0.544 & 0.712 & 0.95 & 0.64 & 0.11 & 0.13 & 0.32 & 0.07 & 0.69 & 0 & 0.64  \\
        & 3D-FCN & 0.735 & 0.563 & 0.723 & 0.97 & 0.46 & 0.27 & 0.44 & 0.53 & 0.11 & 0.36 & 0.2 & 0.77  \\
        & Attention U-Net & 0.738 & 0.596 & 0.751 & 1 & 0.66 & 0 & 0.55 & 0.18 & 0 & 0.02 & 0 & 0.89 \\
		& Swin-Unet & \textbf{0.936} & 0.900 & 0.932 & 1 & 0.97 & 0.52 & 0.83 & 0.73 & 0.16 & 0.93 & 0.09 & 0.77  \\
		\midrule
		\multirow{4}{*}{Siberia} 
        & \textcolor{black}{Random Forest} & \textcolor{black}{0.677} & \textcolor{black}{0.362} & \textcolor{black}{0.634} & \textcolor{black}{0.93} & \textcolor{black}{0.31} & \textcolor{black}{0} & \textcolor{black}{0.34} & \textcolor{black}{0.18} & \textcolor{black}{0.06} & \textcolor{black}{0} & \textcolor{black}{0.18} & \textcolor{black}{0.42} \\
        & 3D-FCN & 0.854 & 0.760 & \textcolor{black}{0.842} & 0.94 & 0.87 & 0 & 0.58 & 0.75 & 0.21 & 0.16 & 0.01 & 0.87 \\
        & Attention U-Net & 0.903 & 0.847 & \textcolor{black}{0.885} & 1 & 0.97 & 0 & 0.42 & 0.61 & 0.56 & 0.06 & 0.01 & 0.99  \\
		& Swin-Unet & \textbf{0.974} & 0.959 & \textcolor{black}{0.974} & 1 & 0.99 & 0 & 0.86 & 0.95 & 0.79 & 0.75 & 0.92 & 0.98 \\
		\bottomrule
	\end{tabular}
	\label{models_results}
	\vspace{5mm}
	\begin{minipage}{\textwidth}
		\centering
		\tiny{\textit{* \textbf{pa}: Producer Accuracy; \textbf{0}: No data, \textbf{1}: Forest, \textbf{2}: Shrubland, \textbf{3}: Grassland, \textbf{4}: Cropland, \textbf{5}: Wetland, \textbf{6}: Bareland, \textbf{7}: Built-up, \textbf{8}: Water}}
	\end{minipage}
\end{table*}

\section{Results}
\label{sec:results}

For each case study, all S1 acquisitions in 2021 were considered, and four subsequences identified according to the seasons. This selection resulted into a collection of 5105 SAR images for Amazonia (1264 winter, 1310 spring, 1338 summer and 1193 autumn data sets), 5.827 SAR images for Africa (1470 winter, 1480 spring, 1654 summer and 1405 autumn data sets), and 3396 SAR images for Siberia (768 winter, 744 spring, 984 summer and 900 autumn data sets). All the images are GRD, IW, VH, descending orbit data sets.

As mentioned in Section \ref{sec:swin_unet_cfg}, for model training, the channel axis of the input tensors to the networks was used to store the spatio-temporal information contained within the sequence of seasonal features extracted from the original SAR images. By doing so, the network accepts an input of $B \times T \times W \times H$ size, where $B$ is the batch size, $T$ is the temporal dimension, while $W$ and $H$ represent the width and the height of the images, respectively.


\begin{figure*}[htp]
	\begin{minipage}[t]{\textwidth}
		\centering
		\caption{Visual comparison between input S1 (first row), Ground Truth (second row) and predicted patches (third) for the Amazonia region.}
		\includegraphics[width=\linewidth]{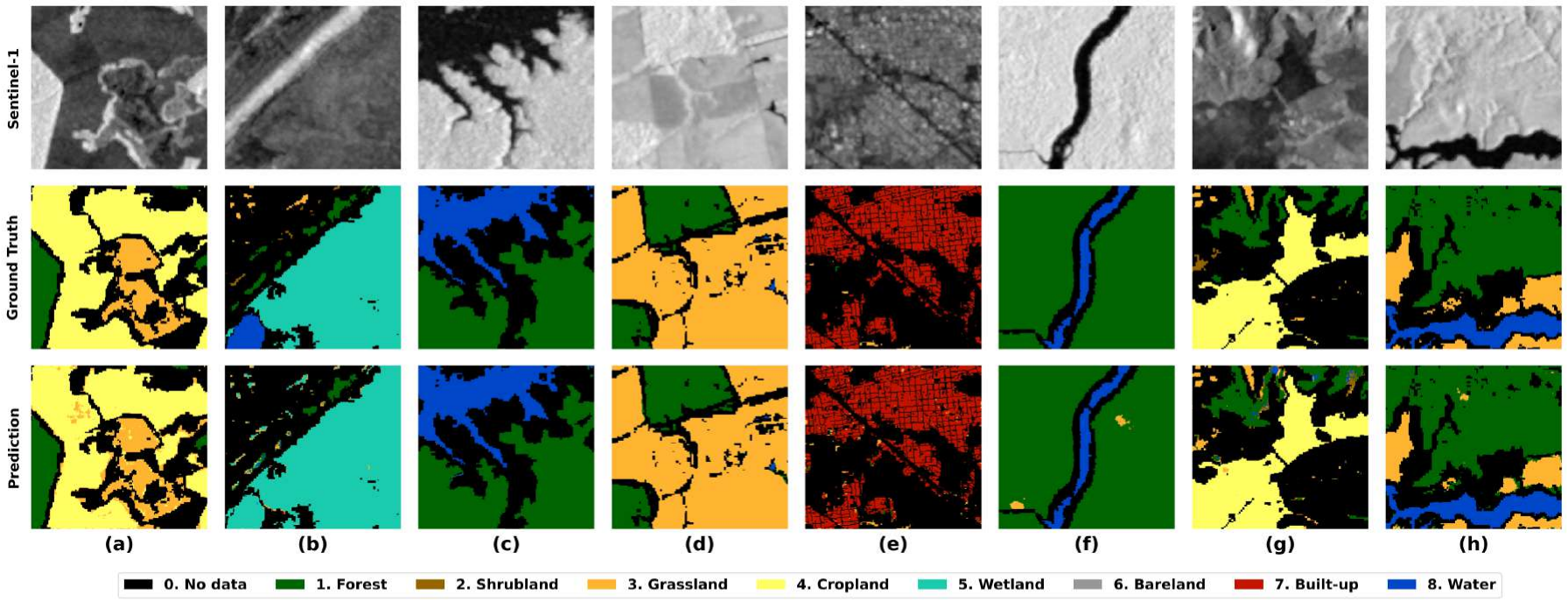}
		\label{final_comparison_Amazonia}
	\end{minipage}
	
	\begin{minipage}[t]{\textwidth}
		\centering
		\caption{Visual comparison between input S1 (first row), Ground Truth (second row) and predicted patches (third row) for the African region.}
		\includegraphics[width=\linewidth]{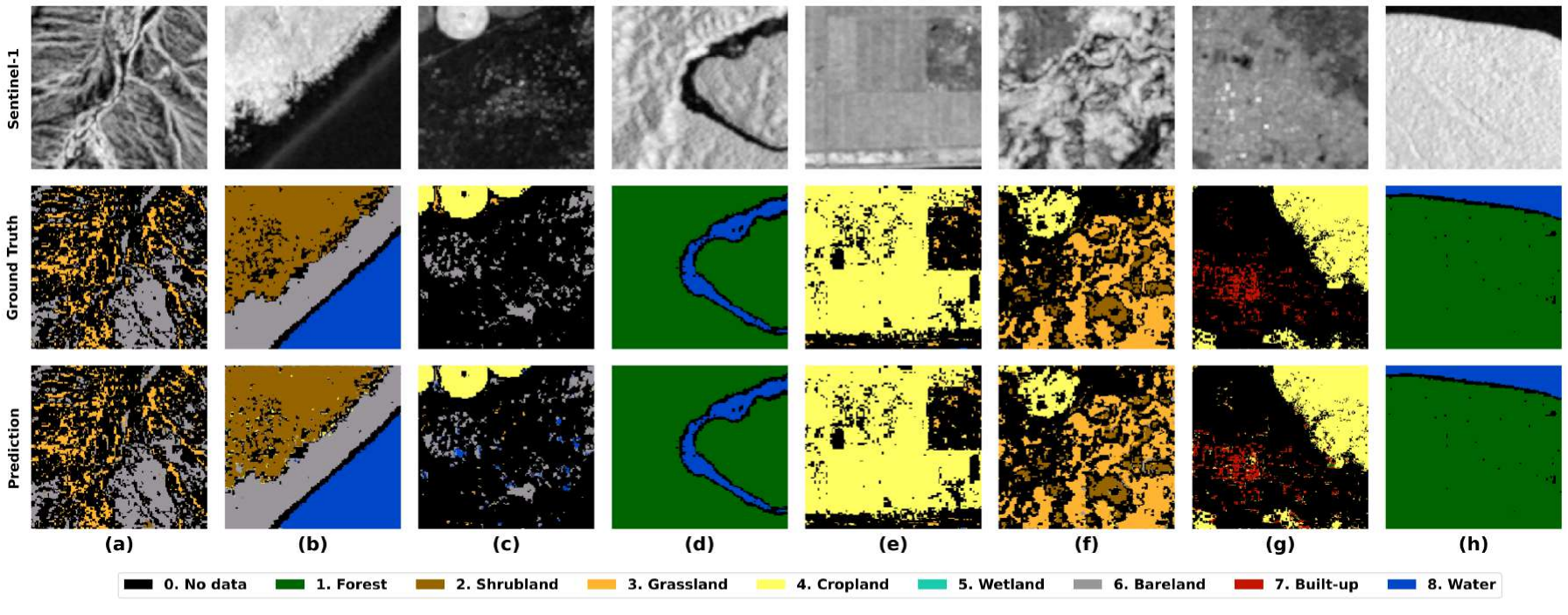}
		\label{final_comparison_africa}
	\end{minipage}
	
	\begin{minipage}[t]{\textwidth}
		\centering
		\caption{Visual comparison between input S1 (first row), Ground Truth (second row) and predicted patches (third row) for the Siberian region.}
		\includegraphics[width=\linewidth]{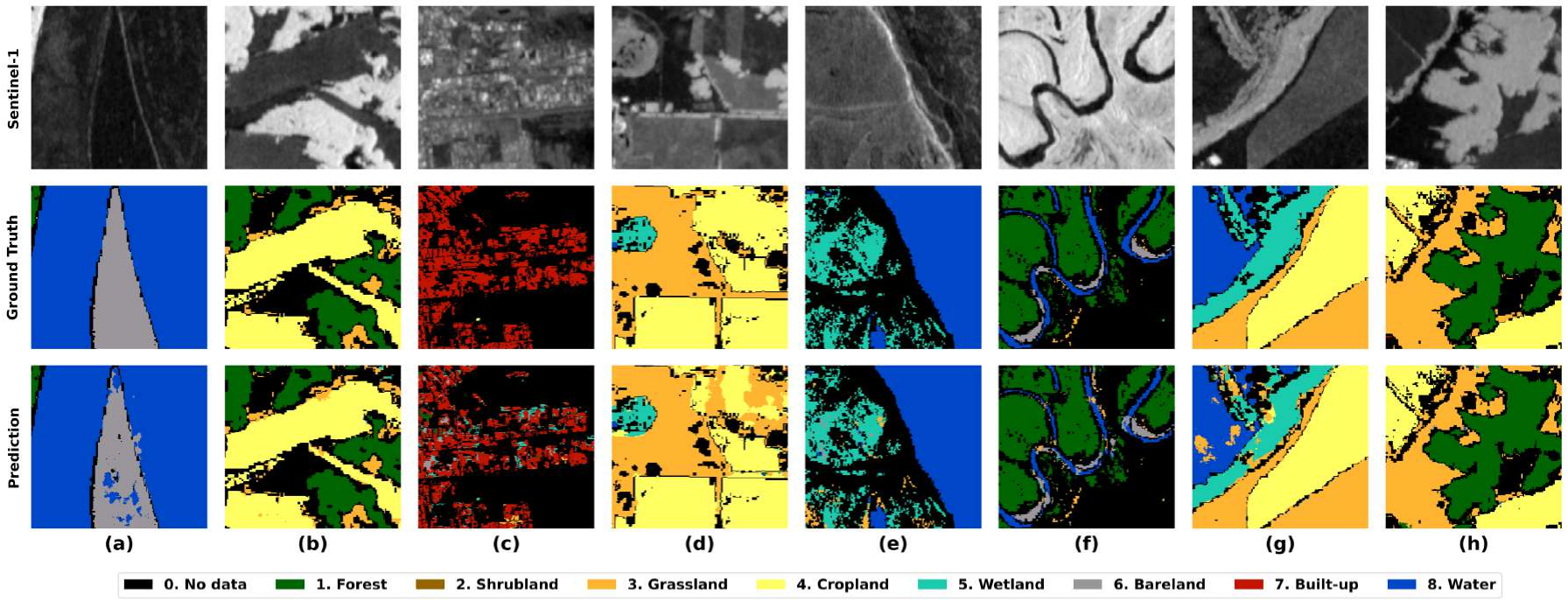}
		\label{final_comparison_siberia}
	\end{minipage}
\end{figure*}

\subsection{Results across different study areas}
\label{subsec:areas_results}

\begin{figure}[htp]
	\centering
	\begin{minipage}{\columnwidth}
		\centering
		\includegraphics[width=0.8\textwidth]{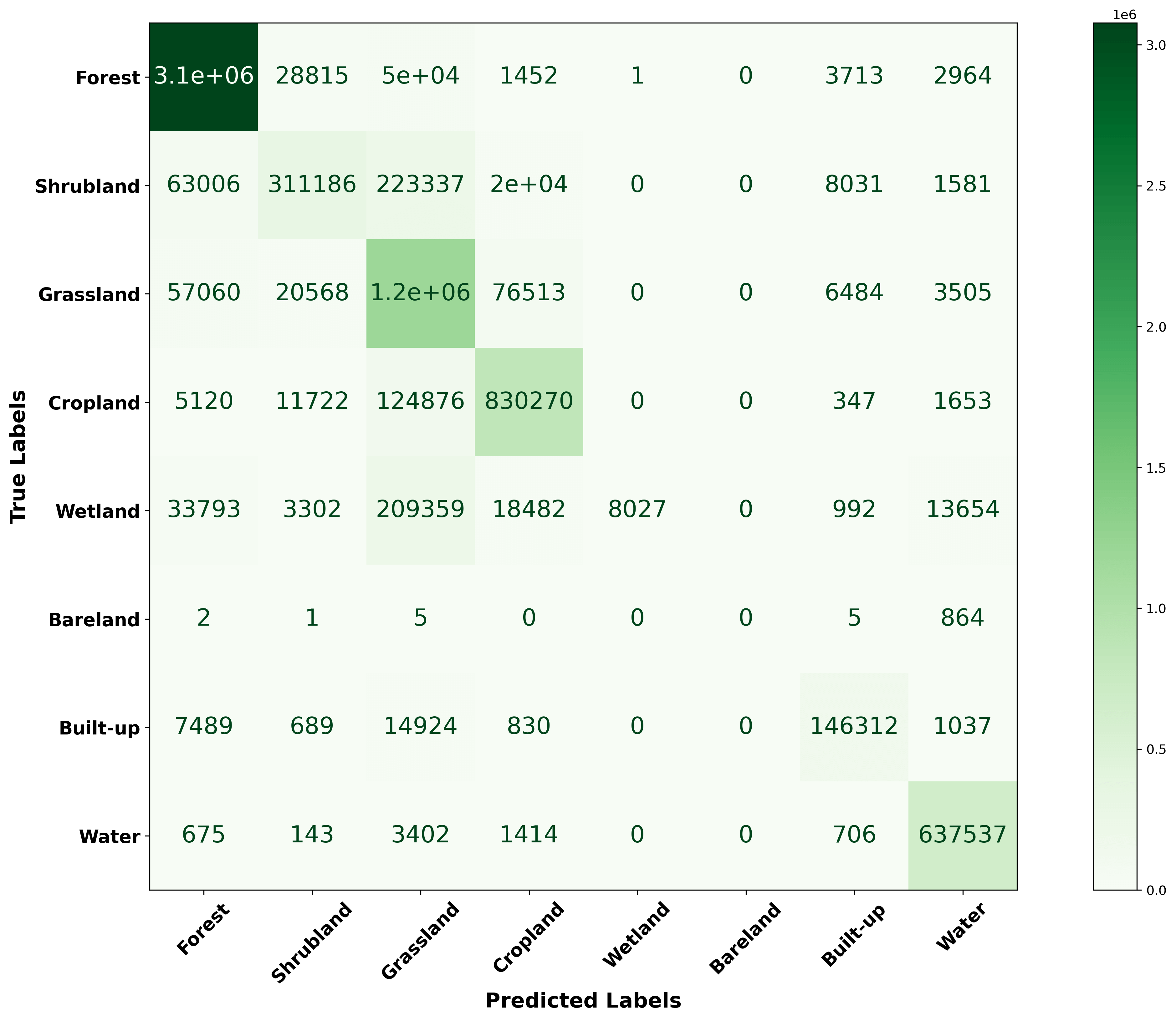}
		\caption{Confusion matrix based on the validation set for the Amazonia region.}
		\label{Amazonia_confmat}
	\end{minipage}
	
	\vspace{\baselineskip} 
	
	\begin{minipage}{\columnwidth}
		\centering
		\includegraphics[width=0.8\textwidth]{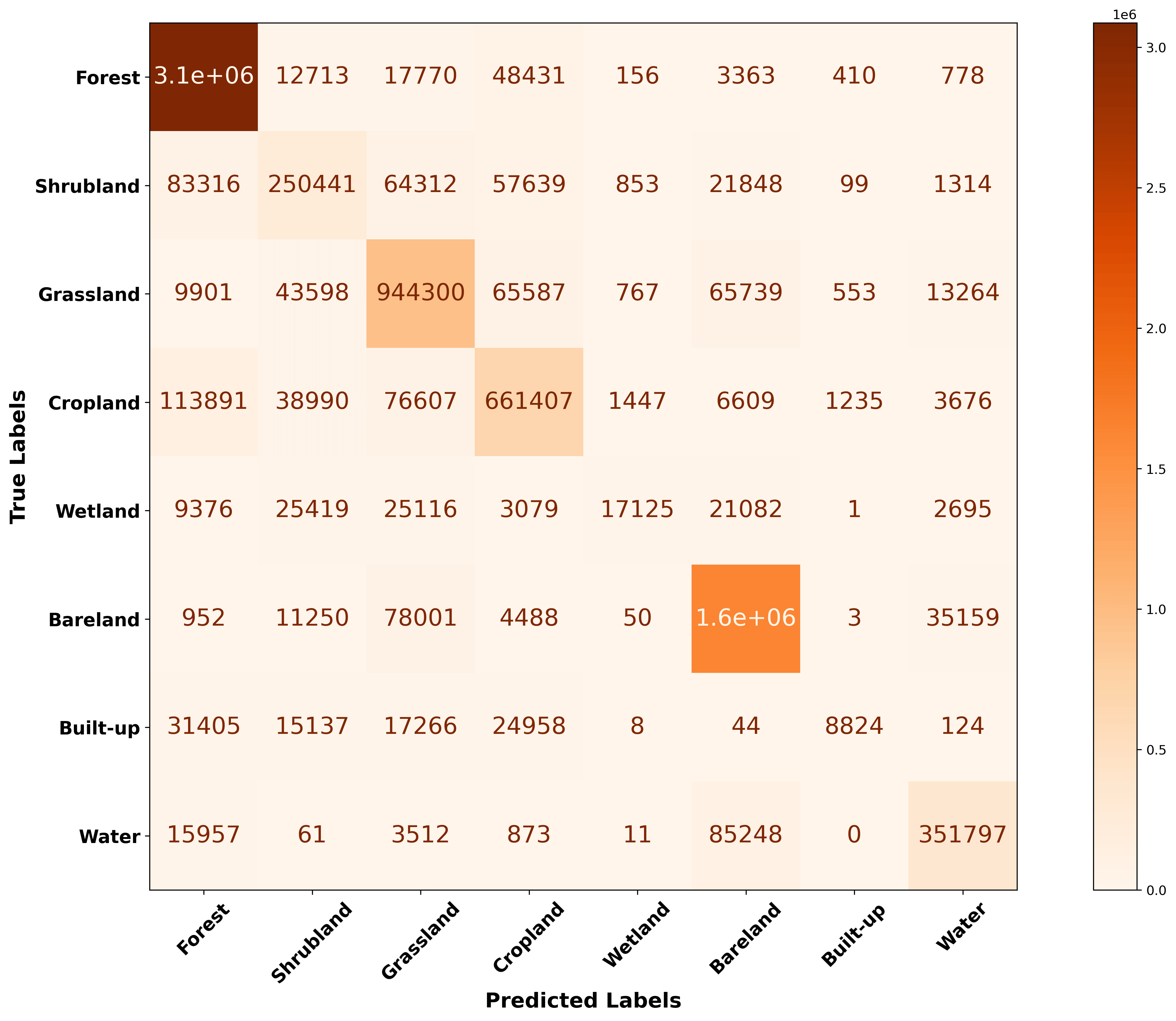}
		\caption{Confusion matrix based on the validation set for the African region.}
		\label{africa_confmat}
	\end{minipage}
    	
	\vspace{\baselineskip} 
	
	\begin{minipage}{\columnwidth}
		\centering
		\includegraphics[width=0.8\textwidth]{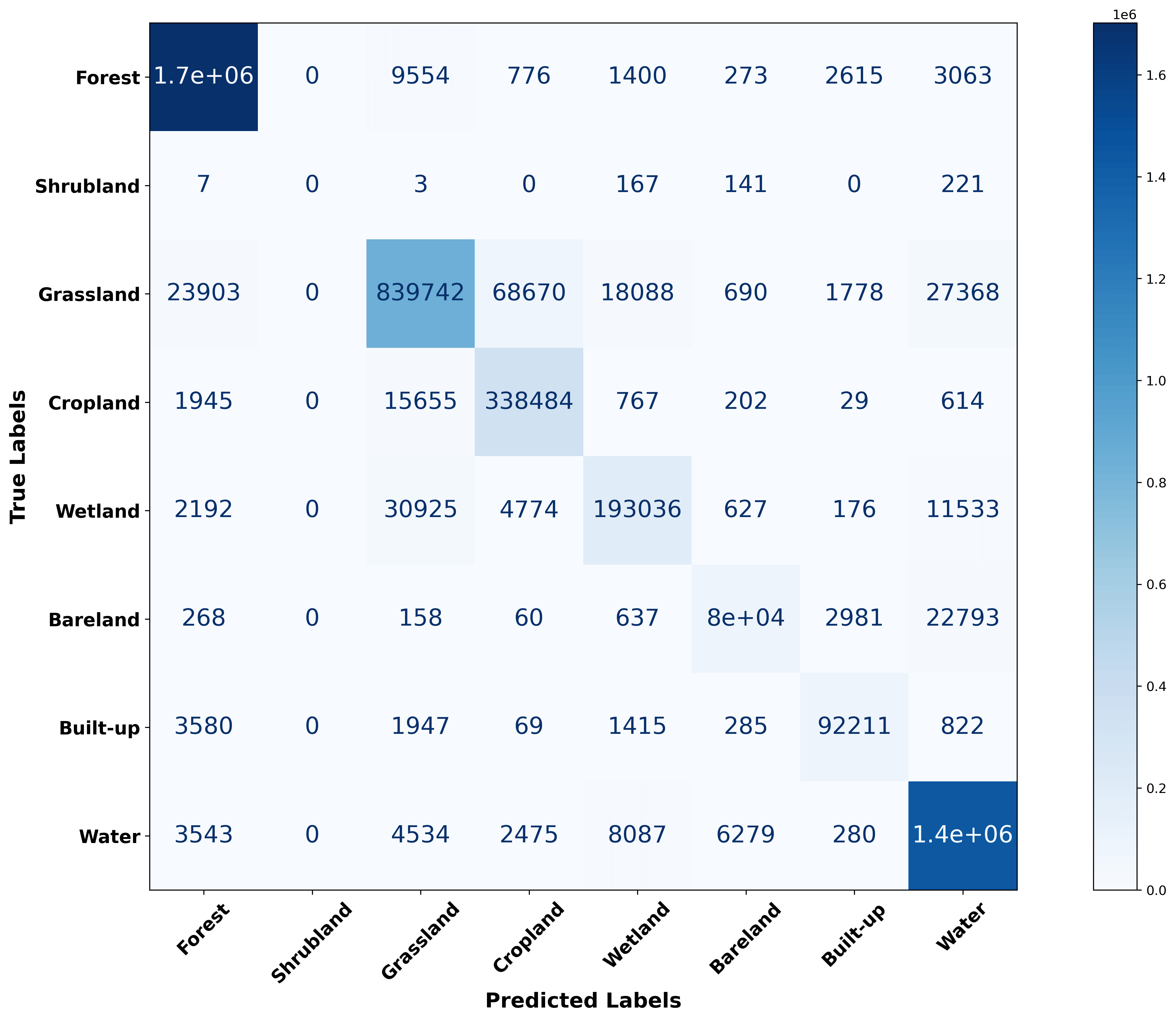}
		\caption{Confusion matrix based on the validation set for the Siberian region.}
		\label{siberia_confmat}
	\end{minipage}
\end{figure}

\begin{figure}[htp]
	\centering
	\includegraphics[width=\columnwidth]{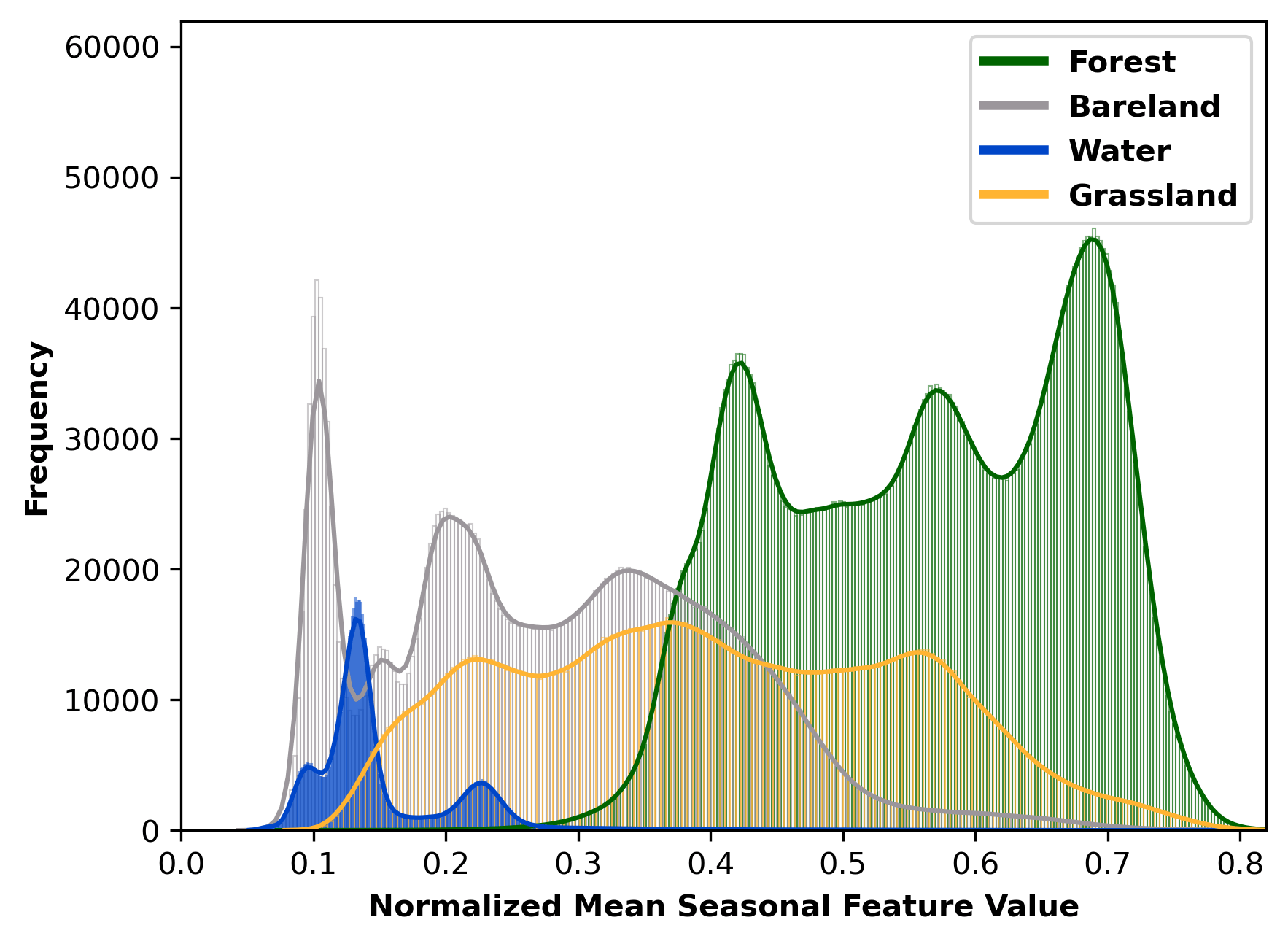}
	\caption{Histogram distribution of the normalized mean values of the stacked seasonal features for each input pixel in the African dataset, corresponding to different LC classes: \textit{Forest}, \textit{Bareland}, \textit{Water}, and \textit{Grassland}, showing the overlap among these classes.}
	\label{histogram_bareland}
\end{figure}

\begin{figure}[t]
	\centering
	\includegraphics[width=\columnwidth]{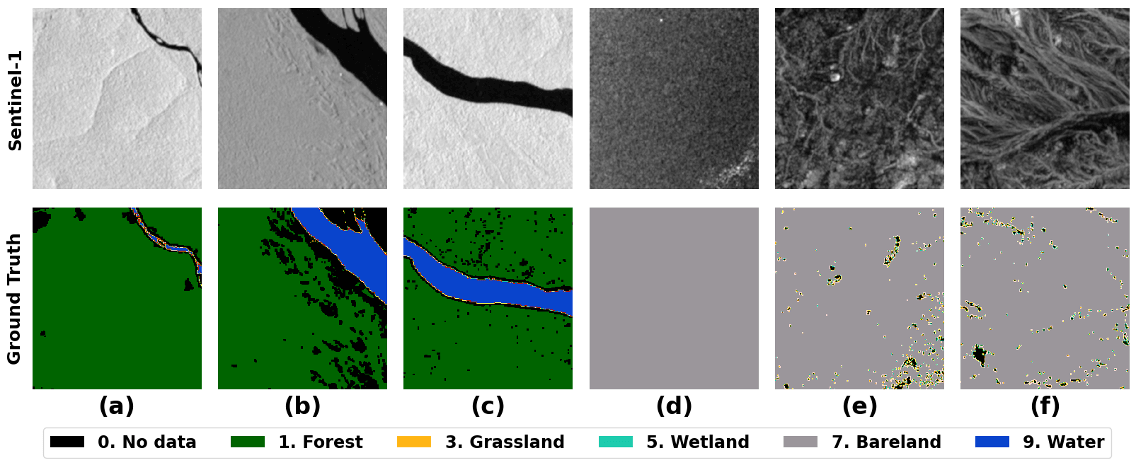}
	\caption{Comparison between the spatial patterns of \textit{Forest} and \textit{Water} (a, b, c) when compared to the more complex spatial pattern of the patches containing the \textit{Bareland} class (d, e, f). }
	\label{bareland_pattern}
\end{figure}

\begin{figure*}[htp]
	\centering
	\includegraphics[width=2\columnwidth]{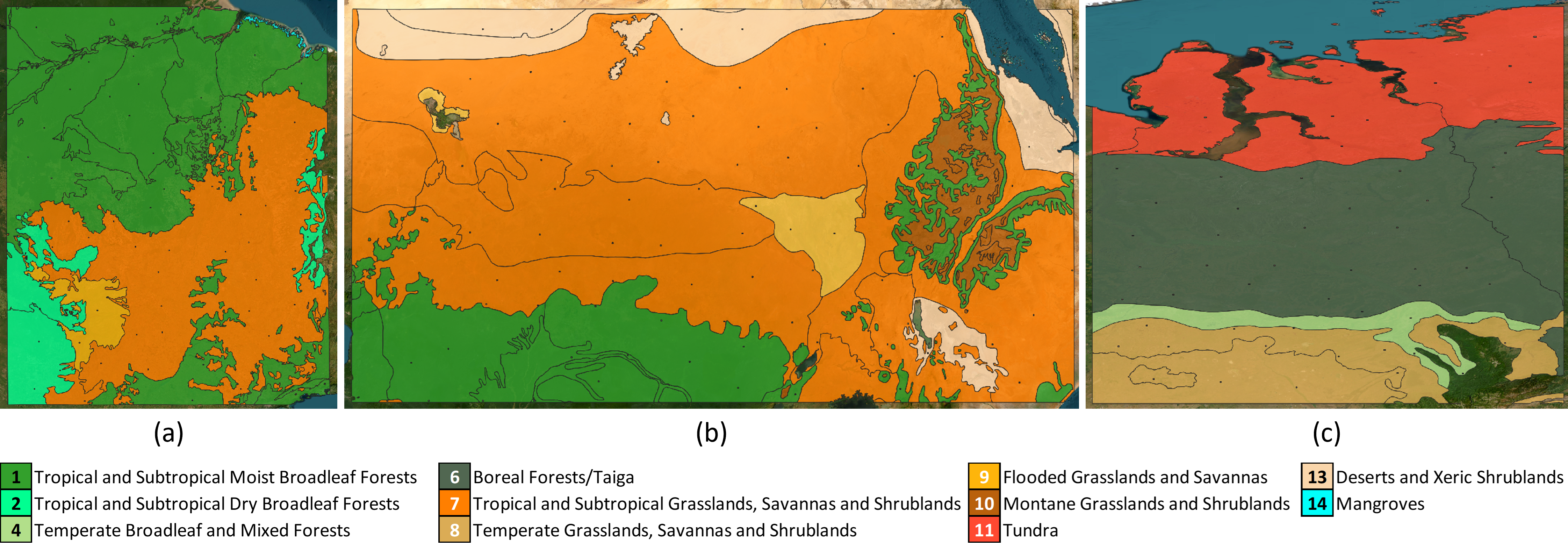}
	\caption{Subdivision of the MOLCA patches of the training set according to the ecoregions for (a) Amazonia, (b) Africa and (c) Siberia. The legend provides the information on the colormap and numerical values used to identify each climatological area.}
	\label{figure:ecoregions}
\end{figure*} 

\begin{figure*}[htp]
	\centering
	\includegraphics[width=2\columnwidth]{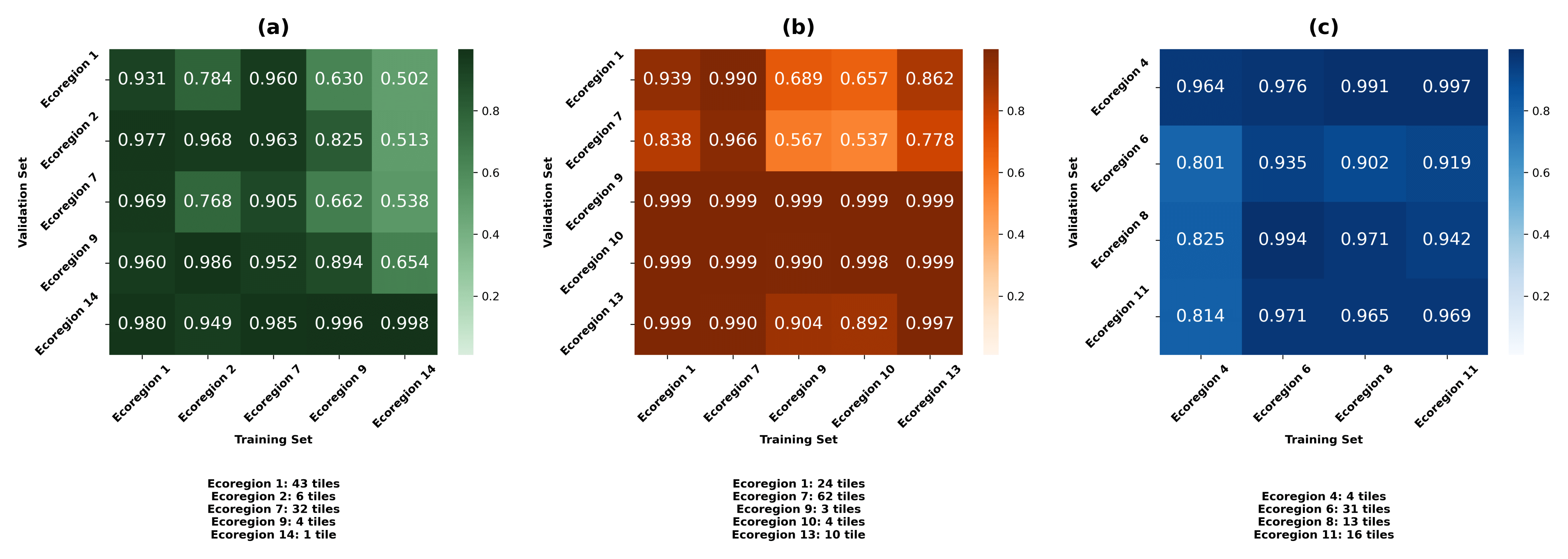}
	\caption{Overall Accuracy (O.A.) matrices obtained according to possible training/test ecoregion combinations for (a) Amazonia, (b) Africa, and (c) Siberia.}
	\label{ecoreg_train_val_am_af_sib}
\end{figure*} 

The results for the three examined regions—Amazonia, Africa, and Siberia—are presented in Table \ref{models_results}, where the outcomes obtained using the best-performing model (\textit{Swin-Unet}) are compared with two reference CNN-based models, \textit{3D-FCN} \cite{GambaMarzi} and \textit{Attention-Unet} \cite{oktay2018attention}. A visual comparison of the results obtained with the \textit{Swin-Unet} model, including random S1 inputs from the validation set, Ground Truth, and corresponding predictions, is depicted in Figures \ref{final_comparison_Amazonia}, \ref{final_comparison_africa}, and \ref{final_comparison_siberia}.
In these areas, the \textit{Swin-Unet} model has achieved O.A.s of 93.3$\%$, 93.6$\%$, and 97.4$\%$, respectively. The associated confusion matrices are illustrated in Figures \ref{Amazonia_confmat}, \ref{africa_confmat}, and \ref{siberia_confmat}, where each element along the diagonal represents the producer's accuracy (PA) of the respective class. The PA associated with the \textit{No data} class has been omitted from the matrix representations since the masking operation explained in Section \ref{proposed_approach} led to a $100\%$ PA for this specific class. It is only reported for completeness in the aforementioned Table \ref{models_results}.

Notably from Table \ref{models_results} \textit{Forest}, \textit{Grassland}, \textit{Cropland} and \textit{Water} classes are very well extracted using the best model in all three cases of the study areas.

The comparison of the utilized models on the African dataset reveals that the CNN-based methods may struggle to distinguish \textit{Bareland} from classes like \textit{Water} or \textit{Grassland} in SAR images. 
The histogram based on the normalized mean values of the stacked seasonal features for each input pixel in the African dataset (Fig. \protect\ref{histogram_bareland}) shows that the Forest class is more clearly separated, while there is overlap among the \textit{Bareland}, \textit{Water}, and \textit{Grassland} classes.

In Africa, the 3D-FCN gives 0.77 for \textit{Water}, but 0.36 for \textit{Bareland}, 0.53 for \textit{Cropland} and 0.44 for \textit{Grassland}. The Attention U-Net framework instead achieved PA of 0.89 for \textit{Water}, 0.66 for \textit{Forest}, and 0.55 for \textit{Grassland}, but only 0.02 for \textit{Bareland}. In contrast, Swin-Unet achieves a final PA of 0.97 for \textit{Forest}, 0.83 for \textit{Grassland}, and 0.93 for \textit{Bareland}, outperforming the other two models.

The enhancement in \textit{Bareland} recognition is evident not only in Africa but also in the Siberian dataset. Despite comprising only $1.04\%$ of the available tiles, \textit{Bareland} achieves a PA of 0.75 through the Swin-Unet model, a significant improvement over the initial 0.16 and 0.06 values obtained with the 3D-FCN and Attention U-Net, respectively.
These results still indicate that the transformer model recognizes various LC classes and tends to make more balanced decisions compared to the other two CNN-based models, which, conversely, demonstrate high PA only for specific classes like \textit{Water}. 
This discrepancy may arise from the Attention U-Net model's limited generalization capability, leading to good classification performance only for classes with easily recognizable spatial patterns or unique brightness values, such as \textit{Water} or \textit{Forest} (Fig. \ref{bareland_pattern} a), b), c)). However, this results in poorer performance when distinguishing the more complex morphological relationships of specific LC classes like \textit{Bareland} (Fig. \ref{bareland_pattern} d), e), f)).
Therefore, Attention U-Net, relying on locality-dependent attention mechanisms, may struggle with classes sharing similar pixel value distributions.

Regarding the 3D-FCN model, its three-dimensional logic-based structure does not offer additional advantages when the input data is not a dense temporal sequence but instead consists of seasonal synthetic images, the so-called \textit{features}. In this context, a 2D-CNN model, like the Attention U-Net, appears to be sufficient. Additionally, the process of masking \textit{No data} values, as previously described in Section \ref{subsection:molca_set}, hampers the learning of continuous spatio-temporal relationships, potentially resulting in a loss of contextual information. In contrast, Transformer-based models like \textit{Swin-Unet} consider global pixel relationships, leading to a more accurate assessment of context and environment, as explained in Section \ref{intro}. The global attention mechanisms in these models can help recover this context, which CNNs cannot achieve due to the local nature of the convolutional kernel.
Additionally, the \textit{Built-up} class showed significant improvement through the Swin-Unet model in Amazonia and Siberia with final values of 0.85 and 0.92.

For the African region, the value for this specific class remains poor, likely due to the lower number of representative labels for the aforementioned class in this area (only 0.5$\%$).

Another contributing factor could be the small and fragmented nature of urban areas, particularly when compared to more spatially uniform LC classes like \textit{Forest}, \textit{Grassland}, or \textit{Cropland}. In favor of these latter classes, the highest confusion is observed when predicting \textit{Built-up}, indicating a significant challenge in visually distinguishing small urban areas from surrounding classes due to their reduced size and scattered appearance.

For \textit{Wetland}, a high PA of 0.79 is obtained in the Siberian region while, for the other two areas, the accuracy remains low. In both cases, this class is often misclassified with \textit{Grassland}, but this is likely due to the fact that samples from this LC class are represented by only $0.8\%$ and $0.6\%$ of the Amazonian and African datasets, respectively.

Finally, for the \textit{Shrubland} class, the best-performing model demonstrates significant improvements in PA, achieving final values of 0.50 (Amazonia) and 0.52 (Africa). In contrast, the Attention U-Net model consistently yields zero accuracy scores for this LC class across all cases, while the 3D-FCN struggles with final values of 0.08 and 0.27, respectively. However, in Siberia, PA remains zero for all three considered DL models, as this class class represents only $0.006\%$ of the region.

It is worth noting that the obtained PA values for the \textit{Shrubland} class in Africa and Amazonia may appear good but not outstanding when analyzed individually. This is due to the specific scenarios in which this class is situated. For instance, as shown in Fig. \ref{final_comparison_africa} b), classification benefits from a noticeable contrast in backscattering values between \textit{Shrubland} and adjacent classes within the S1 image. Conversely, scenarios like those depicted in Fig. \ref{final_comparison_africa} f) may exhibit slightly higher confusion due to the spatial contiguity between the target class and neighboring LC types, such as \textit{Grassland}. As observed in Figures \ref{Amazonia_confmat} and \ref{africa_confmat}, the model tends to confuse \textit{Shrubland} with \textit{Grassland} in $36\%$ and $13\%$ of cases, respectively.\\

The classification performance of three DL architectures was assessed against one of the most widely used non-parametric supervised classifiers, the RF algorithm \cite{ho1995random}, known for its application in LC mapping utilizing both optical and SAR data \cite{amini2022urban}, \cite{monsalve2022evaluation}, \cite{eisavi2015land}.
An analysis of the results presented in Table \ref{models_results} reveals that, for both the Amazon and Siberia, the RF model exhibits lower final O.A.s compared to all three DL approaches. In Africa, the RF model achieves a final O.A. of 0.745, which is marginally higher than the O.A. of both the 3D-FCN and Attention U-Net models. However, the RF underperforms overall, as reflected by its Kappa coefficient of 0.544 and F1-Score of 0.712, both of which are slightly lower than those of the Attention U-Net and 3D-FCN models. Moreover, the RF model is notably outperformed by the Swin U-Net architecture, which achieves significantly higher metrics, with a final O.A. of 0.936, a Kappa of 0.900, and an F1-Score of 0.932.
The classification trend aligns with previous findings from the DL models, demonstrating moderate performance in recognizing the \textit{Forest} and \textit{Water} classes. Specifically, in the Amazon, the final recorded PA is 0.51 for the \textit{Forest} class and 0.18 for the \textit{Water} class. In Africa, the PAs reach 0.64 for both the \textit{Forest} and \textit{Water} classes, while in Siberia the PAs are 0.31 for \textit{Forest} and 0.41 for \textit{Water}.
For other classes, the RF model exhibits significant struggles, particularly with the \textit{Shrubland} (PA = 0.09), \textit{Cropland} (PA=0.07), \textit{Bareland} (PA = 0), and \textit{Built-up} (PA = 0.02) categories in Amazonia, as well as \textit{Shrubland} (PA = 0.11), \textit{Wetland} (PA = 0.07) and \textit{Built-up} (PA=0) in Africa. Similar challenges are evident in Siberia, where PAs for \textit{Shubland} (PA=0), \textit{Wetland} (PA=0.06), and \textit{Bareland} (PA=0) areas are also low. These PA values indicate a marked inability to identify these classes, suggesting that the model may be inadequately trained or that these classes suffer from high intra-class variability, complicating accurate classification.
This analysis underscores the limitations of the RF algorithm in contexts characterized by complex LC types. Research indicates that although RF is effective for various applications, DL approaches generally outperform it when nuanced classification is required due to their capacity to learn hierarchical feature representations \cite{pashaei2020review}.
The effectiveness of DL models in managing complex datasets and capturing intricate patterns in LC classification is evident. In environments like the Amazonia, DL architectures can leverage their ability to learn hierarchical features, resulting in improved classification accuracy. For instance, studies have demonstrated that the Swin U-Net, which employs a shifted window mechanism, excels in capturing both local and global contextual information, thereby significantly enhancing classification performance \cite{li2024deep}.
Moreover, while RF remains a robust option in various scenarios, its effectiveness can be compromised by the high dimensionality and complexity of remote sensing data, particularly in heterogeneous landscapes. This highlights the necessity of employing advanced techniques for LC mapping, where DL methods can better exploit the rich information inherent in the data.

\subsection{Analysis by ecoregions} 
\label{ecoregion_section}

Further experiments were carried out considering the ground truth organized according to ecoregions. Ecoregions are geographic regions of the world that indicate the distribution of ecosystems and plant and animal communities. They have been defined in different ways, depending on the specific purpose of particular regionalization approaches. They are then understood as macro-ecosystems, the largest regional-scale ecosystem units, corresponding to the large climatic regions where climatic conditions are relatively uniform \cite{olson2001terrestrial, bailey2002ecoregion}. Fig. \ref{figure:ecoregions} depicts the MOLCA patches of the training set for Amazonia, Africa and Siberia according to the climatological regions. The legend reports the colormap and the numerical code of each region.

In this section, the experiments were performed by considering these climatic subdivisions within the three considered study areas. This supplementary analysis is designed to assess the results of training and testing within the same ecoregion compared to different ones, aiming to enrich the depth of this evaluation. The results obtained using the \textit{Swin-Unet} model are reported in Figures \ref{ecoreg_train_val_am_af_sib} (a), b), c)) and are expressed in terms of O.A. 
Elements on the diagonal represent the cases of training and testing within the same ecoregion, while off-diagonal elements represent cross-validation between different ecoregions.

In the Amazonia region (Fig. \ref{ecoreg_train_val_am_af_sib}(a)) the O.A. is typically higher when both training and testing are conducted within the same ecoregion. However, an exception is noted when using \textit{Ecoregion 14} and \textit{Ecoregion 9} for training, coupled with testing in other ecoregions. The lower performance in this scenario can be attributed to \textit{Ecoregion 14} and \textit{Ecoregion 9} having only one and four tiles, respectively, which likely limits the diversity and volume of training data.

A similar pattern is observed in the African area (Fig. \ref{ecoreg_train_val_am_af_sib}(b)), where lower performance is particularly noticeable when using \textit{Ecoregion 9} and \textit{Ecoregion 10} for training. These two contain only 3 and 4 tiles, respectively, which may not provide sufficient data for the model to generalize effectively across more diverse ecoregions.

In the Siberian area (Fig. \ref{ecoreg_train_val_am_af_sib}(c)), the O.A. values remain impressively high, exceeding $80\%$ in all possible combinations. It is noteworthy, however, that the lowest values occur when training in \textit{Ecoregion 4} and testing in other ecoregions. Despite this, the accuracy still remains consistent, even with \textit{Ecoregion 4} comprising only 4 tiles. This highlights the robust generalization capabilities of the best performing model in this particular area, demonstrating its effectiveness even with limited data from certain ecoregions.

\subsection{Comparative Analysis with SAR Time Series}
\label{sec:standard_comparative_analysis}

This subsection presents a comparative analysis using standard time series data for the three considered regions to demonstrate the effectiveness of the synthesized seasonal feature extraction approach. The objective is to show that this method yields superior results compared to acquiring multiple time-step images for each season.
To ensure alignment with the synthesized seasonal approach, five separate temporal images were collected per season for each MOLCA tile, culminating in a total of twenty images per tile for the year 2021. This method provided a comprehensive and consistent dataset across regions, totaling 1620 S1 images for Amazonia, 1860 for Africa, and 960 for Siberia. This structured dataset ensures high temporal resolution, enhancing the consistency and reliability of the seasonal analysis for each MOLCA tile within these distinct ecological zones. Images were carefully selected to ensure complete coverage of each tile, encompassing 100$\%$ of the tile area. To assure data consistency and spatial completeness, images failing to meet the required spatial coverage threshold, along with tiles lacking sufficient temporal coverage, were systematically excluded from the dataset. This selective process preserved the quality and uniformity of the dataset, retaining only tiles with robust temporal representation suitable for detailed analysis. As a result, the finalized dataset included 81 MOLCA tiles for Amazonia, down from the initial 86; 93 tiles for Africa, reduced from 103; and 48 tiles for Siberia, compared to an initial 64. The seasonal distribution of the acquired S1 time series is illustrated in Table. \ref{table:time_series}.

Notably, in Amazonia, Africa, and particularly in Siberia, the limited availability of multitemporal data leads to some information loss, as fewer tiles meet the requirement of five acquisitions per season.

\begin{table*}[htp]
	\centering
	\caption{Seasonal count of S1 images from the 2021 time series for Amazonia, Africa, and Siberia}
	\begin{tabular}{lccccccc}
		\toprule
		\textbf{Region} & \textbf{Platform} & \textbf{Acquisition mode} & \textbf{Product format} & \textbf{Polarization} & \textbf{Resolution} & \textbf{Season*} & \textbf{Number of images}\\
		\midrule
		\multirow{4}{*}{Amazonia} 
        & S1A$/$S1B & IW & GRDH & VH & 10 m & Winter & 405 \\
        & S1A$/$S1B & IW & GRDH & VH & 10 m & Spring & 405 \\
        & S1A$/$S1B & IW & GRDH & VH & 10 m & Summer & 405 \\
		& S1A$/$S1B & IW & GRDH & VH & 10 m & Autumn & 405 \\
		\midrule
		\multirow{4}{*}{Africa} 
        & S1A$/$S1B & IW & GRDH & VH & 10 m & Winter & 465 \\
        & S1A$/$S1B & IW & GRDH & VH & 10 m & Spring & 465 \\
        & S1A$/$S1B & IW & GRDH & VH & 10 m & Summer & 465 \\
		& S1A$/$S1B & IW & GRDH & VH & 10 m & Autumn & 465 \\
		\midrule
		\multirow{4}{*}{Siberia} 
        & S1B & IW & GRDH & VH & 10 m & Winter & 240 \\
        & S1B & IW & GRDH & VH & 10 m & Spring & 240 \\
        & S1B & IW & GRDH & VH & 10 m & Summer & 240 \\
		& S1B & IW & GRDH & VH & 10 m & Autumn & 240 \\
		\bottomrule
	\end{tabular}
	\label{table:time_series}
	\vspace{3mm}
	\begin{minipage}{\textwidth}
		\centering
		\tiny{\textit{* \textbf{Winter}: 01.01 to 03.31, \textbf{Spring}: 04.01 to 06.30, \textbf{Summer}: 07.01 to 09.30, \textbf{Autumn}: 10.01 to 12.31}}
	\end{minipage}
\end{table*}

\begin{table*}[htp]
    \centering
    \caption{Results for standard input time series for Amazonia, Africa and Siberia.}
    \resizebox{\textwidth}{!}{ 
        \begin{tabular}{lccccccccccccc}
            \toprule
            \textbf{Region} & \textbf{Model} & \textbf{O.A.} & \textbf{kappa} & \textbf{F1-Score} & $\textbf{pa}_0$ & $\textbf{pa}_1$ & $\textbf{pa}_2$ & $\textbf{pa}_3$ & $\textbf{pa}_4$ & $\textbf{pa}_5$ & $\textbf{pa}_6$ & $\textbf{pa}_7$ & $\textbf{pa}_8$ \\
            \midrule
            \multirow{2}{*}{Amazonia} 
            & Random Forest & 0.612 & 0.309 & 0.553 & 0.91 & 0.49 & 0.09 & 0.21 & 0.08 & 0.18 & 0 & 0.02 & 0.16 \\ 
            & Swin-Unet & 0.903 & 0.855 & 0.901 & 1 & 0.92 & 0.75 & 0.68 & 0.67 & 0.14 & 0 & 0.75 & 0.94 \\ 
            \midrule
            \multirow{2}{*}{Africa} 
            & Random Forest & 0.743 & 0.535 & 0.707 & 0.97 & 0.61 & 0.09 & 0.13 & 0.17 & 0.07 & 0.70 & 0 & 0.44 \\ 
            & Swin-Unet & 0.872 & 0.795 & 0.854 & 1 & 0.95 & 0.03 & 0.68 & 0.16 & 0 & 0.89 & 0.27 & 0.77 \\ 
            \midrule
            \multirow{2}{*}{Siberia} 
            & Random Forest & 0.670 & 0.315 & 0.619 & 0.93 & 0.25 & 0.00 & 0.34 & 0.19 & 0.08 & 0 & 0.19 & 0.30 \\ 
            & Swin-Unet & 0.896 & 0.817 & 0.893 & 1 & 0.93 & 0 & 0.44 & 0.73 & 0.65 & 0.57 & 0.28 & 0.79 \\ 
            \bottomrule
        \end{tabular}
    } 
    \label{models_results_time_series}
    \vspace{2mm}
    \begin{minipage}{\textwidth}
        \centering
        \tiny{\textit{* \textbf{pa}: Producer Accuracy; \textbf{0}: No data, \textbf{1}: Forest, \textbf{2}: Shrubland, \textbf{3}: Grassland, \textbf{4}: Cropland, \textbf{5}: Wetland, \textbf{6}: Bareland, \textbf{7}: Built-up, \textbf{8}: Water}}
    \end{minipage}
\end{table*}

\begin{table*}[htp]
    \centering
    \caption{Overall Accuracy (O.A.) by Swin-Unet model across random states for a robust cross-validation with a consistent 70\%-30\% train-test split for Amazonia, Africa, and Siberia.}
    \scriptsize 
    \resizebox{\textwidth}{!}{ 
        \begin{tabular}{lcccccccccc}
            \toprule
            \textbf{Region} & \multicolumn{10}{c}{\textbf{Random State}} \\
            \midrule
              & \textbf{2} & \textbf{12} & \textbf{2} & \textbf{32} & \textbf{42} & \textbf{52} & \textbf{62} & \textbf{72} & \textbf{82} & \textbf{92} \\
            \cmidrule{2-11}
            Amazonia & 0.901 & 0.907 & 0.918 & 0.924 & 0.920 & 0.911 & 0.910 & 0.910 & 0.933 & 0.915 \\
            \cmidrule{2-11}
            Africa & 0.923 & 0.936 & 0.926 & 0.935 & 0.926 & 0.921 & 0.932 & 0.928 & 0.936 & 0.936 \\
            \cmidrule{2-11}
            Siberia & 0.966 & 0.973 & 0.971 & 0.966 & 0.974 & 0.970 & 0.965 & 0.973 & 0.964 & 0.971 \\
            \bottomrule
        \end{tabular}
    }
    \label{results_random_state}
    \vspace{2mm}
    \begin{minipage}{\textwidth}
        \centering
        \scriptsize{\textit{\\*Train-Test Split: 70\%-30\%.}}
    \end{minipage}
\end{table*}

\begin{table*}[htp]
    \centering
    \caption{\scriptsize Comparison of Frames Per Second (FPS) at varying batch sizes (BS) and model complexity (millions of trainable parameters) for the three DL models tested in this work.}
    \scriptsize 
    \resizebox{\textwidth}{!}{ 
        \begin{tabular}{lcc|cccccc}
            \toprule
            \textbf{Model} & \textbf{Complexity (M)} & \textbf{Metric} & \textbf{BS 1} & \textbf{BS 2} & \textbf{BS 4} & \textbf{BS 8} & \textbf{BS 16} & \textbf{BS 32} \\
            \midrule
            \multirow{2}{*}{3D-FCN} & \multirow{2}{*}{26.48} & FPS & \scriptsize 45.34 & \scriptsize 49.98 & \scriptsize 52.37 & \scriptsize 54.09 & \scriptsize OOM & \scriptsize OOM \\
            & & Speedup* & \scriptsize 1.00 & \scriptsize 1.10 & \scriptsize 1.16 & \scriptsize 1.19 & \scriptsize - & \scriptsize - \\
            \midrule
            \multirow{2}{*}{Attention U-Net} & \multirow{2}{*}{34.89} & FPS & \scriptsize 61.77 & \scriptsize 76.81 & \scriptsize 80.78 & \scriptsize 84.40 & \scriptsize 86.28 & \scriptsize 86.58 \\
            & & Speedup* & \scriptsize 1.00 & \scriptsize 1.24 & \scriptsize 1.31 & \scriptsize 1.37 & \scriptsize 1.40 & \scriptsize 1.40 \\
            \midrule
            \multirow{2}{*}{Swin-Unet} & \multirow{2}{*}{25.16} & FPS & \scriptsize 55.15 & \scriptsize 86.63 & \scriptsize 92.96 & \scriptsize 96.93 & \scriptsize 98.49 & \scriptsize 98.81 \\
            & & Speedup* & \scriptsize 1.00 & \scriptsize 1.57 & \scriptsize 1.69 & \scriptsize 1.76 & \scriptsize 1.79 & \scriptsize 1.79 \\
            \bottomrule
        \end{tabular}
    }
    \label{table:model_fps_comparison}
    \vspace{2mm}
    \begin{minipage}{\textwidth}
        \centering
        \scriptsize{\textit{OOM: Out of Memory; M: Millions of parameters; *Speedup: Relative to BS 1}}
    \end{minipage}
\end{table*}

This analysis identifies two primary limitations:
\begin{itemize}
\item \textbf{higher computational cost}: acquiring multiple images at different time steps increases data storage and processing requirements compared to synthesizing information from extracted seasonal features;
\item \textbf{limited data availability}: in certain regions, limited access to consistent multitemporal data restricts comprehensive analysis capabilities.
\end{itemize}
Moreover, a key advantage of utilizing the synthesized seasonal feature extraction lies in the elimination of any need to impose conditions related to spatial coverage. This is achieved as the seasonal temporal dynamics are effectively condensed through the computation of the “super image,” which integrates seasonal variations into a single, comprehensive image. This approach simplifies the process, ensuring that temporal information is consistently represented without gaps or additional spatial constraints, thereby enhancing both the efficiency and coherence of the dataset.\\
The results of this comparative analysis, utilizing the best performing DL model (Swin-Unet) and the RF standard approach, are shown in Table \ref{models_results_time_series}. Both methods utilize the same training and validation sets, ensuring a fair comparison of their performance metrics. The analysis demonstrates a general decline in performance when using standard time-series images compared to the synthesized seasonal features approach, as indicated in Table \ref{models_results}.

\section{Discussions}
\label{sec:discussions}

The study presented a comprehensive evaluation of three DL models—specifically the Swin-Unet, 3D-FCN, and Attention-Unet—across various geographical regions: Amazonia, Africa, and Siberia. The results, as indicated in Table \ref{models_results}, reveal that the Swin-Unet model consistently outperforms the other two architectures in terms of O.A. The model achieved O.A.s of 93.3$\%$, 93.6$\%$, and 97.4$\%$ for Amazonia, Africa, and Siberia, respectively, demonstrating its robustness in handling SAR image data across diverse ecosystems.

The robustness of the Swin-Unet model is further illustrated in Table \ref{results_random_state}, which provides a comprehensive analysis of classification accuracy across various random states. The random state, a parameter used to control the shuffling and splitting of the dataset, ensures reproducibility by generating consistent train-test splits. Using a fixed train-test split ratio of $70\%-30\%$, the model’s performance was assessed by varying this parameter. The results demonstrate the consistency of the Swin-Unet model and underscore the model's ability to handle variations in training data distribution, which is crucial for real-world applications. 

In addition to its robustness, the Swin-Unet model demonstrates significant computational efficiency, as shown in Table \ref{table:model_fps_comparison}. The performance evaluation of the three DL models presented in this table provides an in-depth analysis of Frames Per Second (FPS) and model complexity at varying inference batch sizes (BS), following the methodology outlined in \cite{KHAN2024105308}. This comparison highlights the computational advantages of the Swin-Unet, which achieves the highest FPS across all batch sizes and demonstrates the best speed-up factors, particularly at higher batch sizes, compared to the Attention U-Net and 3D-FCN models. Simulations and FPS evaluations were conducted on a NVIDIA GeForce RTX $3070$ GPU high-performance workstation, with $8192$ MiB of dedicated memory and $128$ GB of system RAM.

As shown in Table \ref{table:model_fps_comparison}, the FPS increases with the batch size due to GPU parallelization, saturating between BS $16$ and $32$. This trend underscores the importance of batch size in optimizing inference speed, particularly when handling high-dimensional spatio-temporal data such as the 28 seasonal features used in this work. 
Furthermore, the Swin-Unet combines high classification accuracy with the lowest model complexity ($25.16M$ parameters) among the three models tested, while the 3D-FCN model, in contrast, experienced out-of-memory (OOM) issues at higher batch sizes. This combination of robustness, computational efficiency, and lower model complexity reinforces the suitability of Swin-Unet for large-scale LC mapping tasks.

A critical aspect of the study, beyond the robustness and computational efficiency of the Swin-Unet model, was its ability to effectively classify LC classes, particularly the \textit{Forest}, \textit{Grassland}, \textit{Cropland}, and \textit{Water} categories, which were extracted with high accuracy across all study areas. Conversely, the classification of the \textit{Bareland} class proved more challenging for CNN-based methods, which struggled to distinguish it from similar classes such as \textit{Water} and \textit{Grassland}. The overlap in pixel value distributions, highlighted by the histogram analysis, underscores the inherent difficulties in classifying these LC types.
In Africa, the 3D-FCN model exhibited lower performance in recognizing the \textit{Bareland} class, achieving a PA of only 0.36, while the Swin-Unet model significantly improved this to 0.93. The challenges faced by the other models could stem from their reliance on locality-dependent attention mechanisms, which may not adequately capture the complex spatial relationships inherent in certain LC types.
Furthermore, the analysis of the RF algorithm against the DL models indicates a general trend of lower performance by RF, particularly in heterogeneous landscapes like those found in Amazonia and Siberia. Despite achieving comparable results to some DL models in specific scenarios, its O.A. and Kappa coefficient were inferior when compared to the Swin-Unet, which showcased superior capability in extracting intricate patterns from the data.

The investigation into ecoregion-specific performance indicates that the Swin-Unet model performs better when the training and testing datasets share similar ecoregional characteristics, yielding higher O.A.s. This is likely because consistent ecoregional characteristics reduce the variability in LC features and backscatter signatures within the datasets, enabling the model to learn more representative patterns and generalize effectively within the same ecoregion. Smaller ecoregions with limited data pose significant challenges, particularly in Amazonia and Africa, where ecological diversity introduces variability in vegetation types, LC patterns, and climatic conditions. This diversity often results in overlapping backscatter signals in SAR imagery, making it difficult for the model to distinguish between similar LC classes. Additionally, the presence of mixed LC types within small tiles further complicates training and limits the model’s ability to generalize effectively across the region. In Siberia, the model's ability to generalize is more apparent due to the region's relative climatic and ecological uniformity. In conclusion, the analysis of ecoregion-specific performance highlights the importance of considering regional characteristics in remote sensing applications. While the Swin-Unet model demonstrates strong generalization capabilities, especially in Siberia, its performance is constrained by limited training data in smaller ecoregions in Amazonia and Africa. The study reveals the detrimental effect of limited training data on model generalization, especially in ecologically diverse regions. Increasing the number of training tiles in underrepresented ecoregions would likely enhance performance. Future work should focus on enhancing training datasets, employing advanced data augmentation techniques, and developing ecoregion-aware models to improve LC classification in diverse ecological contexts. These improvements could significantly enhance the applicability of remote sensing in addressing global environmental challenges.


The incorporation of ecoregions into this analysis is particularly significant. Ecoregions represent geographically distinct areas characterized by specific climatic, ecological, and biological attributes, making them vital for understanding and mitigating the impacts of climate change \cite{sayre2020assessment}. For climatologists working on ESA’s Climate Change Initiative (CCI+), ecoregions provide a structured framework to evaluate how environmental factors influence LC dynamics. Aligning remote sensing products with ecoregion classifications enables the development of models tailored to the unique environmental conditions and ecological processes of each area. This alignment enhances their accuracy and increases their applicability in region-specific studies and decision-making processes. This approach not only enhances the precision of LC mapping but also facilitates more targeted analyses of climate-related phenomena, such as deforestation, desertification, or wetland degradation. The refined SAR-based products generated by the proposed method hold immense potential as robust tools for monitoring and addressing climate change challenges across diverse ecological zones.

The challenges associated with SAR data are also a significant consideration in this study. SAR data is inherently prone to speckle noise, a granular interference that arises due to the coherent nature of radar signals. This noise can obscure fine details in the imagery, reducing clarity and making accurate LC classification more challenging. This issue is particularly pronounced in heterogeneous regions, where the backscatter signatures of different LC types, such as \textit{Bareland} and \textit{Water}, often overlap, complicating classification tasks. Additionally, variations in acquisition geometry and the high sensitivity of SAR to surface moisture and roughness adds another layer of complexity, as these factors can alter the backscatter signal and create ambiguities in classification. Addressing these challenges requires advanced preprocessing techniques, such as speckle filtering and radiometric normalization, as well as model architectures capable of extracting robust features despite these inherent limitations. Future research should also explore the integration of complementary datasets, such as optical imagery, to mitigate these challenges and enhance the overall performance of SAR-based LC classification. For instance, optical imagery can provide detailed spectral information that complements the structural data captured by SAR, improving differentiation between similar LC types such as \textit{Bareland} and \textit{Water} LC classes. Additionally, combining SAR's all-weather capabilities with optical data's sensitivity to vegetation and soil properties could offer a more comprehensive understanding of dynamic environmental processes.

Additionally, the comparative analysis with SAR time series further highlights the advantages of the synthesized seasonal feature extraction method employed in this study. 
Consolidating seasonal variations into a single “super image” simplified the data processing pipeline, achieving greater classification accuracy. In other words, instead of analyzing images from different seasons individually, they are combined into a single representation that retains relevant temporal information. This allows for more precise results and reduces the computational load required for the classification process.
Overall, the results of this study demonstrate the efficacy of the Swin-Unet model for LC mapping using SAR imagery, particularly in complex ecological regions. The findings suggest a promising avenue for future research, which could focus on integrating additional features and employing advanced modeling techniques to further improve classification performance across diverse environmental contexts.

\section{Conclusion}
\label{sec:conclusions}

This study showcases the potential of DL architectures in effectively mapping LC types using SAR data. The novelty of this study lies in the combination of the transformer-based Swin-Unet model with seasonal synthetic features. This unique approach leverages the advanced capabilities of transformers along with tailored synthesized spatio-temporal images to achieve excellent results in terms of O.A. The results indicate that the proposed approach outperforms traditional CNN-based architectures, especially in regions with diverse environmental and climatic characteristics. Furthermore, leveraging multitemporal SAR data for climate change analysis can provide crucial insights into environmental shifts in sensitive regions, utilizing the synthetic information provided by seasonal features to bridge temporal gaps, particularly in critical regions such as the Siberian zone.
Future research will focus on enhancing model generalization across diverse ecoregions through adaptive or transfer learning techniques. Moreover, refining data preprocessing methods to address quality issues inherent in SAR data is essential, as it could significantly advance the field, leading to more precise and reliable LC mapping outcomes. These future directions hold significant promise for advancing the effectiveness and applicability of DL-based approaches in LC mapping using SAR data at large scale.

\bibliographystyle{IEEEtran.bst}
\bibliography{references}

\vspace{-10mm}
\begin{minipage}[t]{0.47\textwidth}
\begin{IEEEbiography}[{\includegraphics[width=1in,height=1.25in,clip,keepaspectratio]{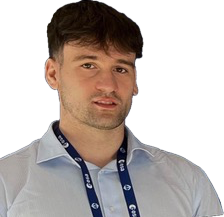}}]{Luigi Russo}
Student Member, IEEE, earned his master’s degree (cum laude) in Electronic Engineering for Automation and Telecommunications from the University of Sannio, Benevento, Italy, in 2023. He is currently pursuing a Ph.D. at the University of Pavia in collaboration with the Italian Space Agency (ASI) in Rome. He has coauthored and presented papers at several prestigious conferences and received the Best Poster Award in Urban and Data Analysis as a young scientist at the 2024 European Space Agency (ESA)-Dragon Symposium. His research focuses on remote sensing and artificial intelligence for the automatic classification and analysis of satellite data, with particular emphasis on creating maps of building exposure and vulnerability to natural hazards and extreme events.
\end{IEEEbiography}
\end{minipage}
\vspace{-3mm}

\begin{minipage}[t]{0.47\textwidth}
\begin{IEEEbiography} [{\includegraphics[width=1in,height=1.25in,clip,keepaspectratio]{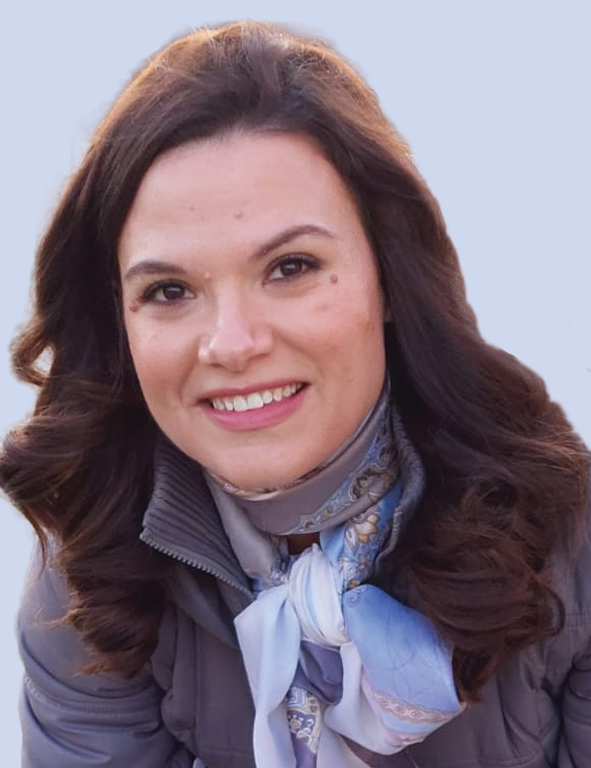}}]{Antonietta Sorriso}
received the M.Sc. degree (summa cum laude) in Telecommunications Engineering and the Ph.D. degree in Information Engineering from the University of Naples "Parthenope", Naples, Italy, in 2015 and 2019, respectively. During her Ph.D., she was a member of the MEG-BioApp Research Unit, Naples, Italy, and her research activities were mainly focused on image and signal processing techniques applied to biomedical imaging, directing her interest towards the analysis of Magnetic Resonance Imaging (MRI) and Magnetoenchephalografy (MEG) data. Since 2019, she is a research fellow at the Department of Electrical, Computer and Biomedical Engineering, University of Pavia, Italy. Her research activities are in the field of SAR image processing, with particular emphasis on denoising and urban detection frameworks.
\end{IEEEbiography} 
\end{minipage}

\vspace{-3mm}

\begin{minipage}[t]{0.47\textwidth}
\begin{IEEEbiography}[{\includegraphics[width=1.05in,height=1.20in,clip,keepaspectratio]{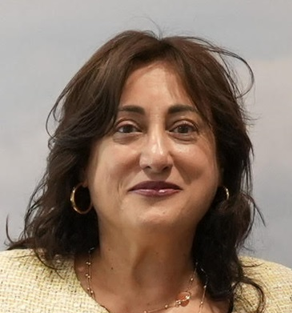}}]{Silvia Liberata Ullo}
IEEE Senior Member, President of IEEE AESS Italy Chapter, Industry Liaison for IEEE Joint ComSoc/VTS Italy Chapter since 2018, National Referent for FIDAPA BPW Italy Science and Technology Task Force (2019-2021). Member of the Image Analysis and Data Fusion Technical Committee (IADF TC) of the IEEE Geoscience and Remote Sensing Society (GRSS) since 2020. GRSS Europe Liaison since January 2024. Editor in Chief of IET IMage Processing. Graduated with laude in 1989 in Electronic Engineering at the University of Naples (Italy), pursued the M.Sc. in Management at MIT (Massachusetts Institute of Technology, USA) in 1992. Researcher and teacher since 2004 at University of Sannio, Benevento (Italy). Member of Academic Senate and PhD Professors’ Board. Courses: Signal theory and elaboration, Telecommunication networks (Bachelor program); Earth monitoring and mission analysis Lab (Master program), Optical and radar remote sensing (Ph.D. program). Authored 90+ research papers, co-authored many book chapters and served as editor of two books. Associate Editor of relevant journals (IEEE JSTARS, IEEE GRSL, MDPI Remote Sensing, IET Image Processing, Springer Arabian Journal of Geosciences and Recent Advances in Computer Science and Communications). Guest Editor of many special issues. Research interests: signal processing, radar systems, sensor networks, smart grids, remote sensing, satellite data analysis, machine learning and quantum ML applied to remote sensing.
\end{IEEEbiography} 
\end{minipage}

\vspace{-3mm}

\begin{minipage}[t]{0.47\textwidth}
\begin{IEEEbiography}[{\includegraphics[width=1in,height=1.15in,clip,keepaspectratio]{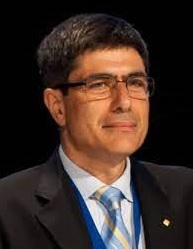}}]{Paolo Gamba}
IEEE Fellow, received the Laurea (cum laude) and Ph.D. degrees in electronic engineering from the University of Pavia, Pavia, Italy, in 1989 and 1993, respectively. He is a Professor of telecommunications with the University of Pavia, where he leads the Telecommunications and Remote Sensing Laboratory and serves as a Deputy Coordinator of the Ph.D. School in Electronics and Computer Science. He has been invited to give keynote lectures and tutorials in several occasions about urban remote sensing, data fusion, EO data, and risk management. Dr. Gamba has served as the Chair for the Data Fusion Committee of the IEEE Geoscience and Remote Sensing Society from 2005 to 2009. He has been elected in the GRSS AdCom since 2014. He was also the GRSS President. He had been the Organizer and Technical Chair of the biennial GRSS/ISPRS Joint Workshops on Remote Sensing and Data Fusion over Urban Areas from 2001 to 2015. He has also served as the Technical Co-Chair of the 2010, 2015, and 2020 IGARSS Conferences, Honolulu, HI, USA, and Milan, Italy, respectively. He was the Editor-in-Chief of the IEEE Geoscience and Remote Sensing Letters from 2009 to 2013.
\end{IEEEbiography}
\end{minipage}

\end{document}